\documentclass[a4paper,fleqn]{cas-dc}

\usepackage[numbers]{natbib}
\usepackage{amsmath}

\def\tsc#1{\csdef{#1}{\textsc{\lowercase{#1}}\xspace}}
\tsc{WGM}
\tsc{QE}
\tsc{EP}
\tsc{PMS}
\tsc{BEC}
\tsc{DE}

\begin{document}
\let\printorcid\relax
\let\WriteBookmarks\relax
\def\floatpagepagefraction{1}
\def\textpagefraction{.001}
\shorttitle{LLM-based task planning for service robots: A review}

\title [mode = title]{Large language model-based task planning for service robots: A review}

\author[1]{Shaohan Bian}

\author[1,2,3]{Ying Zhang}
\ead{yzhang@ysu.edu.cn}
\corref{cor1}
\cortext[cor1]{Corresponding author.}

\affiliation[1]{organization={School of Electrical Engineering, Yanshan University},
                city={Qinhuangdao 066004},
                country={China}}
\affiliation[2]{organization={Engineering Research Center of Intelligent Control System and Intelligent Equipment, Ministry of Education, Yanshan University},
                city={Qinhuangdao 066004},
                country={China}}
\affiliation[3]{organization={Hebei Key Laboratory of Intelligent Rehabilitation and Neuromodulation, Yanshan University},
                city={Qinhuangdao 066004},
                country={China}}

\author[4]{Guohui Tian}
\affiliation[4]{organization={School of Control Science and Engineering, Shandong University},
                city={Jinan 250061}, 
                country={China}}
                
\author[5]{Zhiqiang Miao}
\affiliation[5]{organization={ National Engineering Research Center of Robot Visual Perception and Control Technology, Hunan University},
                city={Changsha 410082},
                country={China}}

\author[6]{Edmond Q. Wu}
\affiliation[6]{organization={Department of Automation, Shanghai Jiao Tong University},
                city={Shanghai 200240},
                country={China}}

\author[7]{Simon X. Yang}
\affiliation[7]{organization={Advanced Robotics and Intelligent Systems Laboratory, School of Engineering, University of Guelph},
                city={Guelph N1G 2W1},
                country={Canada}}

\author[1,2,3]{Changchun Hua}

\begin{abstract}
With the rapid advancement of large language models (LLMs) and robotics, service robots are increasingly becoming an integral part of daily life, offering a wide range of services in complex environments. To deliver these services intelligently and efficiently, robust and accurate task planning capabilities are essential. This paper presents a comprehensive overview of the integration of LLMs into service robotics, with a particular focus on their role in enhancing robotic task planning. First, the development and foundational techniques of LLMs, including pre-training, fine-tuning, retrieval-augmented generation (RAG), and prompt engineering, are reviewed. We then explore the application of LLMs as the cognitive core—“brain”—of service robots, discussing how LLMs contribute to improved autonomy and decision-making. Furthermore, recent advancements in LLM-driven task planning across various input modalities are analyzed, including text, visual, audio, and multimodal inputs. Finally, we summarize key challenges and limitations in current research and propose future directions to advance the task planning capabilities of service robots in complex, unstructured domestic environments. This review aims to serve as a valuable reference for researchers and practitioners in the fields of artificial intelligence and robotics.
\end{abstract}

%


\maketitle

\section{Introduction}\label{Introduction}
With the continuous progress of robotics \cite{bauer2024challenges,zhang2025environment} and artificial intelligence (AI) \cite{dafarra2024icub3,newbury2023deep}, the perception, reasoning, and action capabilities of service robots have been significantly enhanced. As a result, service robots are evolving into capable assistants within domestic environments \cite{xu2024grasp,zhang2022semantic,natarajan2024trust}. By autonomously performing various household tasks such as desktop organizing, floor cleaning and even caring for the elderly, these robots contribute to improving the quality of life \cite{li2025bathing, zhang2020exploring}.

To effectively accomplish daily tasks, service robots must engage in detailed task planning, which involves understanding user requirements, decomposing high-level goals into subtasks, and generating executable action sequences. Task planning is the core technology to realize autonomous decision-making of service robots, which enables the robots to analyze environmental information, evaluate task requirements, and develop optimal action plans. Such autonomous decision-making capability is the key to robot intelligence. Traditional robotic task planning methods mainly rely on detailed task models and environment information \cite{wang2022generalizable,zheng2024knowledge,odense2022neural}. For example, the task planning problem is typically divided into the domain description and problem description \cite{haslum2019introduction}, where the domain describes a set of actions defined by their preconditions and subsequent effects, while the problem description specifies the initial state and the desired goal conditions. However, such approaches generally presuppose finiteness, determinism, and invariance of planning goals. These assumptions limit their applicability, especially in real-world scenarios where environments are dynamic and unpredictable. Moreover, traditional methods often suffer from low fault tolerance, limited scalability, and weak adaptability to real-world interactions.
In real-world domestic environments, the positional relationship between objects and scenes is usually dynamically changing \cite{zhang2022building}. Challenges such as clutter, occlusion between individual objects further lead to an increase in the perception difficulty of service robots \cite{zhao2021hierarchical}. In addition, due to the lack of a priori knowledge of the environment and the target object, the efficiency of target search in large-scale scenarios tends to be low, and blind search of the entire home environment is time-consuming, labor-intensive and difficult to apply to real life \cite{liu2022service, zhang2019efficient}. Therefore, how to generate executable task planning that meets user needs in complex and dynamically changing environments is one of the critical challenges in deploying service robots in unstructured domestic environments.

In recent years, with the continuous development of large language models (LLMs), they have shown great superiority and adaptability in various fields. These giant neural networks with tens of billions of parameters have shown unprecedented comprehension and generation capabilities through training on massive amounts of data. From the stunning performance of ChatGPT to the multimodal breakthrough of GPT-4, LLMs are redefining the boundaries of machine intelligence. The deep fusion of LLM and robotics \cite{sun2024leveraging} is driving the evolution of robots from single-task performers to general-purpose intelligences. With their rich common-sense knowledge and reasoning ability as the cognitive core of the intelligences, these models have dramatically improved the service robot's ability to understand and adapt to the environment. And through the fusion processing of multimodal data such as vision and speech, the robot is able to more accurately perceive and understand the surrounding environment, providing a reliable basis for autonomous decision-making. 
LLMs show even more amazing potential in task planning for service robots. They are able to break down abstract instructions into concrete steps, consider environmental constraints, and formulate reasonable action plans. This ability allows the robots to be no longer limited to preset programs, but to respond flexibly to a variety of complex tasks, understand user needs in depth, and even provide personalized services.

{
		Recent literature has seen a surge in reviews exploring the integration of LLMs with robotics. Kim et al. \cite{kim2024survey} offered a foundational analysis of LLM-enhanced robotic systems, covering communication, perception, and control. Zeng et al. \cite{zeng2023large} and Li et al. \cite{li2025large} delved into specific applications, respectively examining LLMs’ impact on decision-making and path planning, and their versatility in multi-robot coordination. Cui et al. \cite{cui2025task} took a unique task-centric approach, investigating autonomous task discovery through a fusion of visual semantics and LLMs.
		Furthermore, several reviews have focused on narrower, yet critical, subfields. These include surveys on traditional robotics topics like navigation and perception \cite{mavrogiannis2023core,loganathan2023systematic,zhang2023large}, as well as those centered on advanced LLM-driven functionalities such as reasoning and tool use \cite{yang2023foundation,sun2023survey,qin2024tool}.
}

{
		While existing surveys have comprehensively covered various aspects of robotics and LLM integration, our review identifies a critical gap in systematic analysis of task planning specifically for domestic service robots. This domain presents unique challenges due to the unstructured nature of home environments and diverse user interaction patterns, which demand specialized approaches not fully addressed in prior works.
		Unlike previous reviews that broadly survey LLM applications in robotics, our work not only provides a concise and systematic overview of the fundamentals and core techniques of LLMs, emphasizing their application to task planning for service robots operating in home environments, but also pioneers a modality-centric taxonomy (text, vision, audio, multimodal) for analyzing task planning challenges, enabling more targeted technological development. This novel framework reveals previously overlooked intermodal synergies and implementation barriers specific to home service scenarios.
}

{
		The key innovations and contributions of this paper are as follows:
}
\begin{itemize}
\item {
	{
		We present a systematic review of the development of LLMs and provide a concise overview of their core techniques, which include pre-training, fine-tuning, retrieval-augmented generation, and prompt engineering.
}}
\item {
	
		We establish the first modality-centric taxonomy for analyzing LLM-based task planning in domestic service robotics, thereby revealing unique cross-modal challenges.
}

\item {
	
		We critically examine the key obstacles hindering real-world deployment of LLM-based service robots and propose actionable research pathways to advance the field.
}

\end{itemize}

The rest of the paper is organized as follows. In Section \ref{Theory Foundation of LLM}, we first introduce the important theoretical foundations of the LLM, and in Section \ref{Taxonomy} we provide a systematic categorization of LLM-based task planning for home service robots based on the differences in the input modalities, and provide a detailed overview of each category in Sections \ref{Text-based LLM Task Planning} to \ref{Multi-modal Large Language Models Planning}. In Section \ref{Current Challenges and Future Prospects}, we present some challenges to the current LLM-based robotic task planning and an outlook to the future. Finally, we conclude the paper in Section \ref{Conclusion}.

\section{Theory Foundation of LLM}\label{Theory Foundation of LLM}
In this section, we provide an overview of the theoretical foundations of LLM, covering the developmental background of LLM, the key techniques - Pre-training, Fine-tuning, RAG, and Prompt Engineering. In addition, we provide an overview of the application of LLM to robotic task planning, showing how it empowers robots to achieve more efficient and intelligent task execution.

\subsection{Background}
\begin{figure*}[tp]
	\centering
	\includegraphics[width=1.0\textwidth]{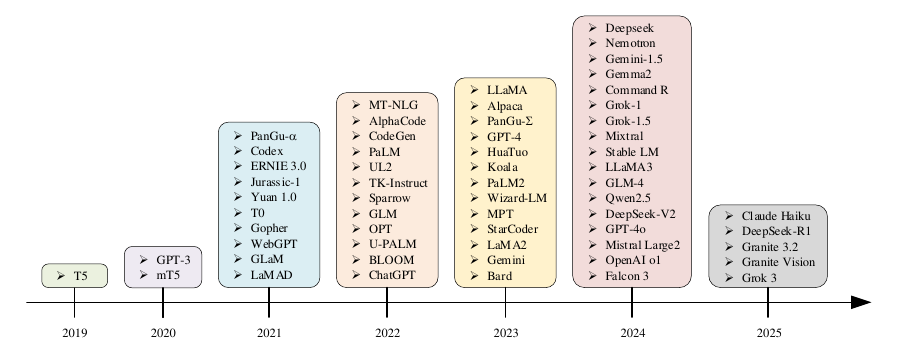}
	\caption{The evolution of LLM since 2019}
	\label{fig:background}
\end{figure*}

Since the proposal of the Transform architecture, the development of LLM has gone through two core breakthrough phases, which have driven the vigorous development of the natural language processing (NLP) field. The first phase is marked by the proposal of Transformer \cite{vaswani2017attention}, whose self-attention mechanism enables parallelized long sequence modeling. GPT-1 \cite{radford2018improving} and \cite{devlin2019bert} adopt unidirectional and bidirectional pre-training, respectively, to lay down the two paradigms of generation and comprehension, and the pre-training-fine-tuning framework achieves significant improvement on several NLP tasks. The second phase focuses on model scale-up and interaction capabilities, with GPT-3 \cite{brown2020language} and ChatGPT \cite{ouyang2022training} demonstrating the potential of generative AI and optimized dialog interactions, respectively. Current research trends are evolving towards efficient, autonomous intelligences and ethical alignment, gradually approaching the boundary of general-purpose artificial intelligence (AGI). Fig. \ref{fig:background} illustrates the evolution of LLMs since 2019, and we describe some of these typical LLMs below.

\textbf{T5} \cite{raffel2020exploring} improves on the traditional Transformer model by moving layer normalization outside of residual connections. It employs masked language modeling as a pre-training target, using spanwise masking of consecutive tokens instead of independent masking of individual tokens, which shortens sequence length and accelerates training. Once pre-training is complete, an adapter layer \cite{houlsby2019parameter} is used to fine-tune for downstream tasks.

\textbf{GPT-3} \cite{brown2020language} continues the architecture of GPT-2 \cite{radford2018improving} and draws on the design of Sparse Transformers \cite{child2019generating}, combining dense and sparse attention mechanisms in the transformer layer and employing a gradient noise scale \cite{mccandlish2018empirical} during training.GPT-3 extends the model parameters to 175 billion, further validating the positive correlation between model size and performance improvement.

\textbf{GLaM} \cite{du2022glam} represents a family of language models employing a sparsely activated Mixed (MoE) structure of decoder experts \cite{fedus2022switch}. GLaM sparsely activates the experts, with each input token being processed by only the optimal two experts. The model keeps the training energy consumption at one-third of GPT-3 while maintaining high performance.

\textbf{GLM-130B} \cite{zeng2022glm} is a bilingual (English-Chinese) model, which is distinguished from the unidirectional GPT-3 \cite{brown2020language} by the use of autoregressive masks to populate the pre-training targets for training and the application of gradient contraction of the embedding layer to ensure training stability.

\textbf{DeepSeek-v2} \cite{liu2024deepseek} is a MoE model that introduces Multihead Latent Attention (MLA) to reduce the inference cost by compressing the key-value (KV) cache into potential vectors. Due to MLA, the inference throughput of DeepSeek-v2 is 5.76 times faster than DeepSeek \cite{bi2024deepseek}.

\begin{table*}[tp]
	\centering
	\caption{Comparison of typical LLMs}
	\label{tab:llm_comparison}
	\begin{tabular}{>{\centering\arraybackslash}m{2.5cm} >{\centering\arraybackslash}m{1.5cm} m{5.8cm} m{5.8cm}}
		\toprule
		\multicolumn{1}{c}{\textbf{Model}} & 
		\multicolumn{1}{c}{\textbf{Parameters}} &
		\multicolumn{1}{c}{\textbf{Advantages}} & 
		\multicolumn{1}{c}{\textbf{Application Areas}}\\ 
		\midrule
		
		T5 \cite{raffel2020exploring} & 11B & Moves layer normalization outside of residual connections, uses spanwise masking, and employs an adapter layer for fine-tuning. & Text classification, question answering, summarization.  \\
		\addlinespace
		GPT-3 \cite{brown2020language} & 175B & Combines dense and sparse attention mechanisms, employs a gradient noise scale, and scales up to 175 billion parameters. & Text generation, language understanding, code generation.   \\
		\addlinespace
		GLaM \cite{du2022glam} & 1200B & Employs a sparsely activated MoE structure, processes each input token by only the optimal two experts, and reduces training energy consumption. & Text generation, language understanding. \\
		\addlinespace
		GLM-130B \cite{zeng2022glm} & 130B & Uses autoregressive masks for pre-training targets and applies gradient contraction of the embedding layer for training stability. & Bilingual text generation (English-Chinese), language understanding.  \\
		\addlinespace
		DeepSeek-v2 \cite{liu2024deepseek} & 236B & Introduces Multihead Latent Attention (MLA) to compress the key-value (KV) cache into potential vectors, significantly reducing inference cost. & Efficient text generation, language understanding. \\
		\addlinespace
		\bottomrule
	\end{tabular}
\end{table*}

\subsection{Pre-training}
The pre-training of LLM is the most critical basic stage in the whole model construction process, and this process is essentially a self-supervised learning approach that allows the model to automatically extract linguistic laws, construct a knowledge system, and form a deep understanding of natural language from massive unlabeled text data. Modern LLM pre-training is mainly built on the Transformer architecture \cite{zhang2023cross}, and according to different application scenarios and functional requirements, researchers have developed three main architectural variants: causal decoder architecture, encoder-decoder architecture, and prefix decoder architecture. Although these architectures differ in their specific implementations, they all rely on Transformer's core component, the self-attention mechanism, to achieve effective modeling of long-distance dependencies through techniques such as multi-attention, residual connectivity, and layer normalization.

Pre-training task design is the core innovation of LLM pre-training. Autoregressive language modeling (e.g., GPT) requires the model to predict the next token based on the above, which perfectly fits the text generation and is the classic pre-training paradigm. In addition, improvement schemes such as continuous fragment masking used by SpanBERT \cite{joshi2020spanbert} and unified language modeling used by UniLM \cite{dong2019unified} enhance the pre-training effect to varying degrees. In addition the model training phase faces the challenge of optimizing the parameters and data on an ultra-large scale. Researchers have developed optimizers such as AdamW \cite{zhou2024towards} and AdaFactor \cite{shazeer2018adafactor} for large-scale training. The combined use of data parallelism, model parallelism, and pipeline parallelism substantially improves the training efficiency. Techniques such as mixed-precision training and gradient checkpointing effectively reduce the video memory occupation, making the training process more efficient.

\subsection{Fine-tuning}
The advantages of fine-tuning are its efficiency and performance improvement. Compared to training from scratch, fine-tuning greatly saves computational resources and time. Pre-trained LLMs are good at predicting textual tokens, but may be limited in generating structured output or processing domain-specific information. To overcome these limitations, the output layer of the LLM can be tuned by Fine-Tuning to fit specific task requirements. To address the lack of knowledge in LLM, fine-tuning training using domain-specific data can enhance LLM's understanding and processing in the domain and improve its accuracy and performance on tasks in the domain.

From the perspective of technical implementation, fine-tuning is mainly divided into two major directions, Full Fine-tuning (FFT) \cite{lv2023full} and Parameter-Efficient Fine-tuning (PEFT) \cite{liu2022few}. FFT updates all the parameters of the model, which is effective but requires a large amount of computational resources and training data, in contrast, PEFT \cite{hu2022lora,houlsby2019parameter,li2021prefix} drastically reduces the resource requirements while guaranteeing the performance by adjusting only some of the parameters or by adding small adaptation modules. In addition, Instruction Fine-tuning \cite{zhang2023instruction} is an important fine-tuning paradigm that has emerged in recent years, which shapes the behavioral patterns of a model by providing it with explicit examples of instructions. Reinforcement Learning with Human Feedback (RLHF) \cite{kirk2023understanding} further enhances the effectiveness of Instruction Fine-tuning by introducing an artificial scoring mechanism to make the model outputs more consistent with human values and preferences.

\subsection{Prompt Engineering}
Prompt Engineering is a key technique in LLM applications that does not require updating the model's parameters and aims to guide the model to generate more accurate and reliable outputs by carefully designing input prompts. A well designed prompt can significantly improve model performance, reduce illusions, and adapt to different task requirements. In the following, we will discuss commonly used prompt methods.

Instruction Prompting \cite{mishra2021reframing} aims to provide the LLM with examples of instruction prompts so that it can eliminate training or testing discrepancies and simulate real-world usage scenarios for chatbots. Zero-shot Prompting \cite{kojima2022large} involves feeding tasks into the model without any examples indicating the desired output, requiring the LLM to answer the user's questions without showing any examples in the prompts. Few-shot Prompting \cite{brown2020language} works by providing the model with a small number of high-quality examples that include the inputs and desired outputs of the target task. By observing these good examples, the model can better understand human intent and the criteria for generating accurate output. Chain-of-Thought Prompting \cite{wei2022chain} generates a series of short sentences, known as chains of reasoning. These sentences describe step-by-step reasoning logic that leads to a final answer, which can be more beneficial for tasks with complex reasoning and larger models. Recursive Prompting \cite{yang2022re3} is a problem solving methodology that involves breaking down complex problems into smaller, more manageable subproblems and then solving these subproblems recursively through a series of prompts.

\subsection{Retrieval-augmented Generation}
RAG [65] is an important technological breakthrough in the field of natural language processing in recent years, which effectively solves the inherent limitations of traditional LLM in terms of knowledge timeliness, factual accuracy, and interpretability by organically integrating the information retrieval system with the generative capability of LLM. The core innovation of the technology lies in the establishment of a dynamic knowledge injection mechanism that enables the generative model to access and utilize external knowledge bases in real time, thus significantly enhancing the reliability and expertise of the generated content.

RAG usually consists of two main components, Retrieval and Generation. First, the model retrieves a set of relevant documents from a pre-constructed document collection based on an input query. The retrieved documents are then fed into the LLM along with the query, and the model generates the final answer based on these inputs. Some typical RAG approaches include Multi-Head RAG (MRAG) \cite{besta2024multi}, which utilizes multi-head attention to enhance multifaceted problem processing. Adaptive-RAG \cite{jeong2024adaptive}, which dynamically selects strategies to cope with queries of varying complexity. Blended RAG \cite{sawarkar2024blended}, which fuses semantic search and hybrid query strategies to improve accuracy. Self-RAG \cite{asai2023self}, enhancing model quality and factuality through search and self-reflection. As well as IM-RAG \cite{yang2024rag}, learning introspective monologue integrates LLM and IR systems to support multi-round RAG. These approaches have their own characteristics, and together they promote the development of RAG technology.

\subsection{Application of LLMs in Robotic Task Planning}

\begin{figure*}[thp]
	\centering
	\includegraphics[width=1.5\columnwidth]{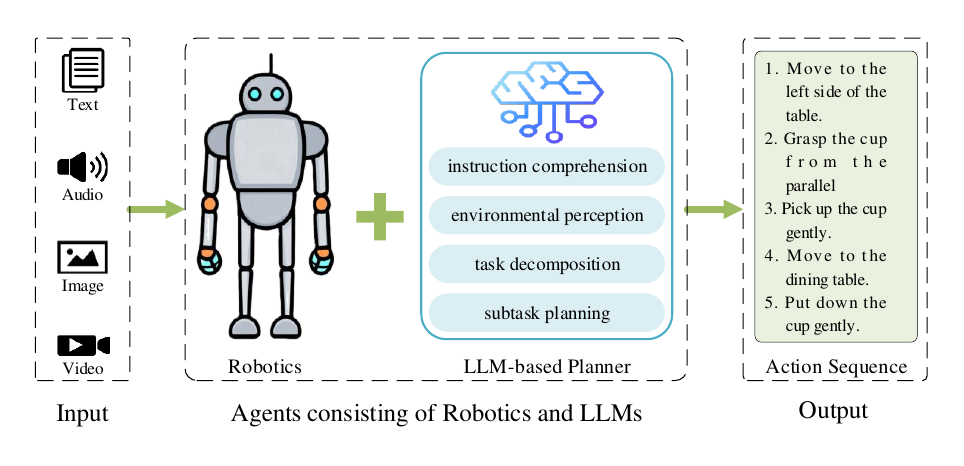}
	\caption{LLM-based task planning for service robots. The robot acquires information in the form of text, video, audio, etc., and passes the processed input signals to the LLM, which carries out the planning of detailed action sequences through natural language processing, task decomposition, and so on.}
	\label{fig:application}
\end{figure*}

In recent years, LLMs have made significant progress in the field of natural language processing. With the continuous evolution of technology, the application of LLMs has broken through the traditional scope of text generation and analysis, and begun to penetrate into emerging fields such as robot task planning. As shown in Fig. \ref{fig:application}, the combination of LLMs and home service robots is driving the development of more intelligent autonomous agents, with LLMs acting as the “brain” of the service robots, which significantly improves their ability in task planning.


{
		When the robot receives ambiguous commands from the user, LLM is able to parse and understand them to accurately capture the user's intent \cite{ding2023task,singh2023progprompt,silver2024generalized}. For example, in a home service robot scenario, the user may say ``put that thing on the high shelf", and LLM understands that ``that thing" refers to the water cup on the table, and recognizes that ``high" refers to the top shelf of the bookshelf. and recognizes that ``high up" refers to the top shelf of the bookcase. Based on this, it combines logical reasoning to determine the optimal execution plan, such as picking up the water cup first, then planning a path to reach the bookshelf avoiding obstacles, and finally placing the water cup at the specified location.	
}

{
		For better task planning, the robot may also rely on VLMs to enhance environment perception \cite{li2024fine,wu2023embodied}. In the above home service scenario, the VLM can recognize the exact location of the water cup, the shape and height of the bookshelf, as well as the obstacles (e.g., chairs or carpets) on the path, providing precise physical information for task planning. Faced with a complex task, the robot is able to break it down into a series of executable sub-tasks, with detailed, robot-executable task planning for each sub-task. For example, a robot can break down ``get a glass of water" into steps such as ``move to the table" and ``grab the glass". The steps of ``navigating to the bookshelf", ``adjusting the posture", and ``placing the cup" are broken down for ``placing the cup". In this process, LLMs assisted in detailed planning to ensure that each subtask was properly executed, thus improving the efficiency and accuracy of the overall task planning.
}

\section{Taxonomy}\label{Taxonomy}

\begin{figure}
	\centering
	\includegraphics[width=.9\columnwidth]{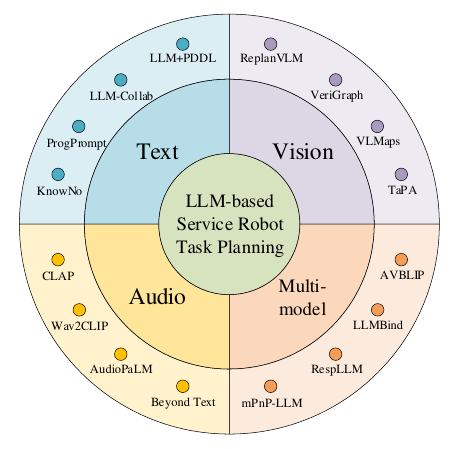}
	\caption{LLM-based task planning taxonomy for service robots. From the point of view of different input modalities, it is categorized into four categories such as text input, visual input, audio input, and multimodal input.}
	\label{fig:Taxonomy}
\end{figure}

Against the backdrop of the intelligent development of domestic service robots, task planning systems are evolving from unimodal to multimodal fusion. Based on the differences in input modalities, this paper constructs a systematic classification framework, dividing LLM-driven task planning methods for domestic service robots into four major categories: Text-based LLM Planning, Vision-Language Model Planning, Audio-based Planning, and Multimodal Large Language Models Planning, as illustrated in Fig. \ref{fig:Taxonomy}. This framework elucidates the technical characteristics and performance boundaries under different modality combinations. Here, we briefly summarize these four categories as follows.

\textbf{Text-based LLM Task Planning.} As the most fundamental implementation paradigm, this planning method primarily relies on text-only input for task planning. Such systems typically employ Transformer-based language models as core processors, utilizing natural language understanding (NLU) modules to parse user instructions and generate corresponding action sequences. However, due to the lack of direct environmental perception capabilities, these systems require pre-constructed semantic maps or knowledge graphs to supplement environmental information.

{
		\textbf{Vision-based LLM Task Planning.} This method significantly enhances robots' understanding of physical environments by integrating visual perception capabilities. VLMs focus primarily on establishing semantic alignment between visual and linguistic modalities.  VLMs usually adopt dual-encoder architectures or cross-modal Transformers to realize efficient visual-linguistic interactions, and their typical applications include:VQA, image description generation,visual localization.
		Through cross-modal attention mechanisms, the system establishes a joint "vision-language" representation space, enabling complex instructions requiring spatial reasoning, such as "place the mug on the dining table into the dishwasher".
}

\textbf{Audio-based LLM Task Planning.} The system enables the robot to understand and respond to voice commands by introducing an audio processing module. Such systems usually include speech recognition module and speech synthesis module, which can convert user 's voice commands into text and generate corresponding action sequences. Through the analysis of audio signals, robots can recognize and respond to specific voice commands in noisy environments, thereby improving their flexibility and adaptability in practical applications.

{
		\textbf{Multimodal LLM Task Planning.} The Multimodal LLM (MLLM) cover not only vision and speech, but also integrate audio, sensor data, robot state, and other modalities to support more complex task planning. MLLMs have three main features: multimodal inputs, dynamic modal fusion, and cross-modal reasoning. Such systems usually adopt a hierarchical fusion strategy: modal alignment is achieved through cross-attention at the feature level, and the contribution weight of each mode is dynamically adjusted by the gating mechanism at the decision level. In this way, the MLLM Planning system can integrate multiple information sources in a complex and changing home environment to generate more accurate and flexible task planning.
}

\section{Text-based LLM Task Planning} \label{Text-based LLM Task Planning}

As the basic paradigm of LLM-driven task planning, text-based systems establish a key baseline for human-computer interaction through natural language processing. These systems convert unstructured natural language instructions into executable action sequences by using the semantic understanding ability of the converter architecture, representing the most computationally efficient method in the modal spectrum.

The task planning system of LLM service robot based on text faces two core challenges: one is how to effectively deal with instructions of different complexity (from simple short-term tasks to complex long-term planning), and the other is how to make up for the lack of environmental perception caused by pure text input. To this end, this section will systematically elaborate the existing research results from two dimensions: the planning method based on instruction complexity and the compensation technology of environment-aware defects.

\subsection{Planning method based on instruction complexity}

	\begin{table*}[h]
	\centering
	\caption{Typical Literature on Text-based LLM Task Planning}
	\label{tab1:llm_planning}
	\begin{tabular}{>{\centering\arraybackslash}m{2cm} >{\centering\arraybackslash}m{1.5cm} m{5.8cm} m{5.8cm}}
		\toprule
		\multicolumn{1}{c}{\textbf{Paper}} & \multicolumn{1}{c}{\textbf{Year}} & \multicolumn{1}{c}{\textbf{Core Innovation}} & \multicolumn{1}{c}{\textbf{Limitations}}\\
		\midrule
		
		\makecell{Ding et al. \cite{ding2023task}} & \makecell{2023} & 
		
		Achieves cross-environment object rearrangement by combining semantic knowledge with geometric planning & 
		Requires precise object labeling and limited to tabletop-scale tasks\\ 
		
		\addlinespace
		
		\makecell{Singh et al. \cite{singh2023progprompt}} & \makecell{2023} & 
		Enables generalized planning through programmatic prompts supporting multi-modal specifications & 
		Demands complex prompt engineering with high computational overhead \\ 
		
		\addlinespace
		
		\makecell{Silver et al. \cite{silver2024generalized}} & \makecell{2024} & 
		Ensures plan validity through direct PDDL integration with formal correctness guarantees & 
		Depends on pre-defined PDDL domains and scales poorly to novel objects \\ 
		
		\addlinespace
		
		\makecell{Ren et al. \cite{ren2023robots}} & \makecell{2023} & 
		Provides statistical uncertainty quantification while minimizing human intervention & 
		Tends to produce conservative plans with complex calibration requirements \\ 
		
		\addlinespace
		
		\makecell{Wei et al. \cite{wei2022chain}} & \makecell{2022} & 
		Simplifies complex reasoning via chain-of-thought prompting without model modification & 
		Lacks formal verification and sensitive to prompt phrasing variations \\ 
		
		\addlinespace
		
		\makecell{Liu et al. \cite{liu2024delta}} & \makecell{2024} & 
		Enables long-horizon planning through autoregressive goal decomposition & 
		Suffers from error accumulation in sub-tasks requiring fine-tuning \\ 
		
		\addlinespace
		
		\makecell{Kannan et al. \cite{kannan2024smart}} & \makecell{2024} & 
		 Facilitates multi-robot collaboration by handling capability heterogeneity	& 
		 Requires centralized coordination introducing communication overhead \\ 
		
		\bottomrule
	\end{tabular}
\end{table*}

\begin{figure}
	\centering
	\includegraphics[width=.9\columnwidth]{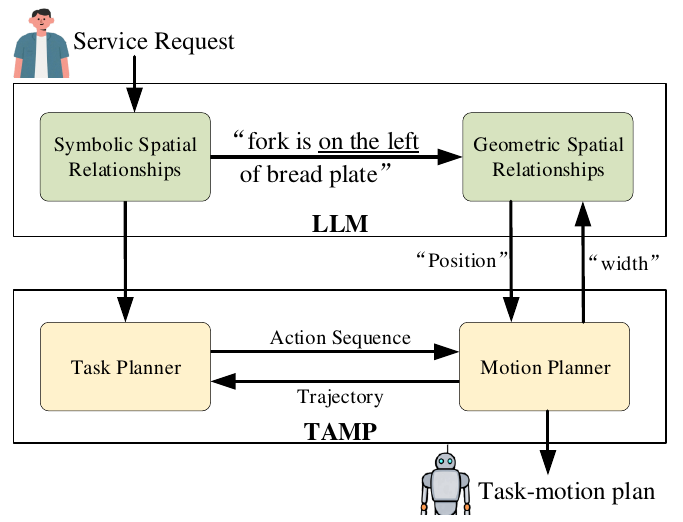}
	\caption{Schematic diagram of the architecture of LLM-grop.}
	\label{fig:text1}
\end{figure}

Due to the excellent performance of LLM in generalization and common sense reasoning, it can skillfully understand natural language instructions and perform logical reasoning, translation and zero sample planning. This section will introduce Text-based LLM Planning in terms of the complexity of instructions, and some of the representative work is shown in Table \ref{tab1:llm_planning}. In the face of simple short-term tasks, LLM shows excellent planning ability. The LLM-grop proposed by Ding et al. \cite{ding2023task} extracts common sense knowledge about semantically valid object configuration from LLM through the prompt function, and instantiates them with tasks and motion planners in order to promote them to different scene geometries, as shown in Figure \ref{fig:text1}. LLM-grop can achieve object rearrangement from natural language commands to human alignment in various environments. ProgPrompt, proposed by Singh et al. \cite{singh2023progprompt}, is a programmatic LLM prompt structure that utilizes the available operations in the environment and the specification of similar programs for objects, so that the plan generation can span the environment, robot capabilities, and tasks. In addition, LLM is often combined with the Planning Domain Definition Language (PDDL) to improve the planning performance of the LLM planner. Sliver et al. \cite{silver2024generalized} directly provided the planning problem in the PDDL syntax to LLM to generate action sequences. Although the planning performance has been improved, this method requires additional knowledge to build a PDDL file. Liu et al. \cite{liu2023llm+} skillfully used the translation function of LLM to convert natural language into PDDL problem, and then solved it by traditional planner, and finally translated the solution back to natural language. Similarly, Tomoya et al. \cite{kawabe2024task} transform natural language tasks into symbolic sequence representations by using LLM, and then derive the best task program by performing task planning based on Monte Carlo Tree Search (MCTS). 

{
		 In the realm of complex long-horizon planning, researchers have explored several strategies to enhance the capabilities of LLM-based robotic systems. One approach involves hierarchical and iterative frameworks. For instance, Chen et al. \cite{chen2025extendable} proposed the Hierarchical Multiscale Diffuser (HM-Diffuser), which employs a hierarchical structure to train on trajectories extended at multiple temporal scales. This is complemented by Progressive Trajectory Extension (PTE), an augmentation method that iteratively generates longer trajectories by stitching shorter ones, thereby efficiently managing tasks across different time horizons.
		 Another direction is the integration of explicit planning modules. Erdogan et al. \cite{erdogan2025plan} introduced the Plan-and-Act framework, which decouples the planning process into two distinct components: a Planner model that generates structured, high-level plans to achieve user goals, and an Executor model that translates these abstract plans into environment-specific actions. To improve the quality of these plans, their framework also incorporates a scalable synthetic data generation method.
		 Furthermore, a widely adopted strategy to mitigate the complexity of long-horizon tasks is hierarchical task decomposition. This approach breaks down a complex mission into a sequence of manageable sub-goals. For example, Wei et al. \cite{wei2022chain} leveraged chain-of-thought (CoT) prompting to elicit step-by-step reasoning, enabling the model to tackle complex tasks more effectively. Building on this, Liu et al. \cite{liu2024delta} proposed DELTA, which decomposes long-horizon objectives into an autoregressive sequence of sub-goals for automated planners. Similarly, Cao et al. \cite{cao2024llm} developed the LLM-Collab framework, which systematically divides complex planning into four stages-analysis, planning, verification, and improvement-each further decomposed into sub-tasks. Zhen et al. \cite{zhen2023robot} merged human expertise with LLMs using a specialized “Think Net Prompt” and a hierarchical decomposition strategy, while Kannan et al. \cite{kannan2024smart} employed LLMs to decompose tasks for multi-robot collaboration, addressing the limitations of single-function robots.
}

However, the possibility of hallucinations in the output of LLM will accumulate due to the increase of planning period. Therefore, in complex long-horizon task planning, how to suppress or reduce the generation of hallucinations is a major difficulty in using LLM for task planning. Ren et al. \cite{ren2023robots} proposed KnowNo, a framework for measuring and adjusting the uncertainty of LLM-based planners. KnowNo is based on conformal prediction theory and provides statistical assurance for mission completion while minimizing manual help in complex multi-step planning settings. Park et al. \cite{park2023clara} designed a LLM uncertainty estimation method to classify whether the command is certain or uncertain. After determining the uncertainty, it classifies the input as certain or uncertain according to a predefined threshold. Once the command is classified as an uncertain command, the LLM generation problem is used to interact with the user to eliminate the ambiguity of the command. Ong et al. \cite{ong2024simple} consider the uncertainty in planning by combining a simple method, which emphasizes quantifying uncertainty and exploring alternative paths for task execution. By setting an appropriate probability threshold in skill selection, a method of measuring uncertainty is established to select a better path to perform tasks. Wang et al. \cite{wang2024llm} proposed a task planning method based on constrained LLM prompt scheme, which can generate executable action sequences from commands. Furthermore, a special processing module is proposed to deal with the LLM illusion problem, so as to ensure that the results generated by LLM are acceptable in the current environment.

\begin{figure}
	\centering
	\includegraphics[width=1.0\columnwidth]{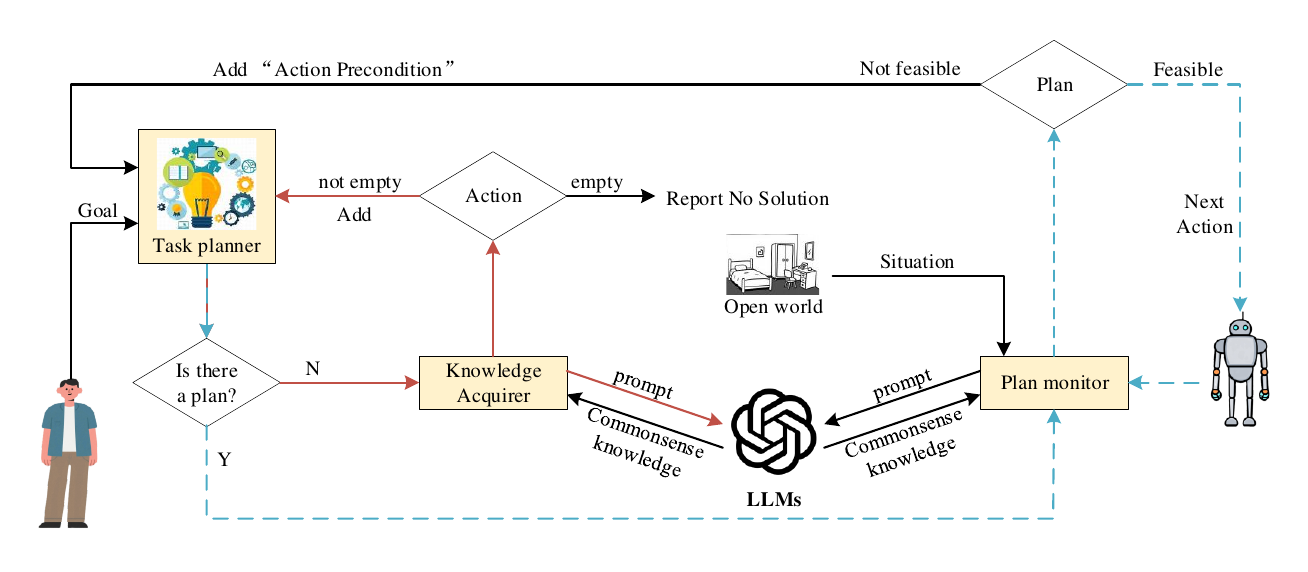}
	\caption{Schematic diagram of the architecture of COWP in paper.}
	\label{fig:text2}
\end{figure}
\subsection{Compensation technology of environmental perception defects}

Although the text-based LLM task planning method has made great progress, it still has an inherent limitation, that is, the perception of the physical environment. The input form of pure text makes researchers have to study compensation strategies to deal with this challenge. The current research mainly solves this problem through static knowledge injection.

Static knowledge injection is shown in LLM-grop \cite{ding2023task}. Before the task planning of desktop object placement, the current environmental information, including known items such as knives, forks, dinner plates, and their existing states, is pre-entered into LLM.Combined with the common sense knowledge of LLM in the form of knowledge base, task planning is performed instead of environmental awareness. Mu et al. \cite{mu2023kggpt} introduced KGGPT, an innovative system that integrates prior knowledge. This system operates by extracting pertinent information from a knowledge graph, transforming it into semantic representations, and then linking these to ChatGPT. KGGPT leverages knowledge graphs to encode robotic skills, task regulations, and environmental limitations. By doing so, it effectively bridges the gap between ChatGPT's existing knowledge base and the practical requirements of real-world service environments. This task planning assumes the invariance and stability of the state of each object in the task space, but in a family environment where dynamic changes may occur, a predefined knowledge base often results in inaccurate or unexecutable task planning results. Dynamic state update better handles the situation where the state of the object may change in an open home environment. Ding et al. \cite{ding2023integrating} developed COWP, an open-world robot planning system that ingeniously integrates a traditional knowledge-based task planning framework with a pre-trained language model, as shown in Figure \ref{fig:text2}. The latter serves as a powerful means for acquiring domain-agnostic common-sense knowledge. This synergy endows COWP with the unique ability to apply universal common-sense knowledge to specialized task-planning scenarios. As a result, COWP can effectively handle unanticipated situations that robots may face during the planning process, enhancing their adaptability in open-ended and dynamic environments. However, this ability to encounter and deal with emergencies is precisely what the large language model of plain text input cannot achieve.

\section{Vision-based LLM Task Planning}\label{Vision-based LLM Task Planning}
The inherent limitations of text-based LLM planners in spatial perception and dynamic adaptation have promoted the rise of VLMs and regarded them as a transformative method in the field of robot task planning. By constructing a unified visual-linguistic representation space, VLM effectively bridges the perception-reasoning gap in a pure LLM system \cite{zhang2025zisvfm}. In the context of service robot task planning, the VLM-based method presents a diversified development trend. In addition, the introduction of visual perception has also stimulated researchers' exploration in active task cognition. In this section, we will explore the research progress of VLM-based task planning methods and active task cognition.

\subsection{Methodological and Frameworks Innovations}
Integrating VLMs into the field of service robot task planning is a major improvement on the traditional text-based method. This part deeply studies the methods and frameworks of using VLMs to enhance the spatial perception and dynamic adaptability of service robots. By bridging the gap between visual perception and language understanding, VLMs enable robots to interpret complex environmental cues and perform tasks with higher accuracy and flexibility. This section will introduce Text-based LLM Planning in terms of the complexity of instructions, and some of the representative work is shown in Table \ref{tab2:vlm_planning}.

\begin{table*}[t!]
	\centering
	\caption{Typical Literature on Methodological and Frameworks Innovations}
	\label{tab2:vlm_planning}
	\begin{tabular}{>{\centering\arraybackslash}m{2cm} >{\centering\arraybackslash}m{1.5cm} m{5.8cm} m{5.8cm}}
		\toprule
		\multicolumn{1}{c}{\textbf{Paper}} & \multicolumn{1}{c}{\textbf{Year}} & \multicolumn{1}{c}{\textbf{Core Innovation}} & \multicolumn{1}{c}{\textbf{Limitations}}\\
		\midrule
		
		\makecell{Li et al. \cite{li2024fine}} & \makecell{2024} & 
		
		Aligns visual attributes with LLM via VLM for precise plans, enhanced by ontological knowledge  & 
		Requires full attribute labeling and predefined ontologies; struggles with novel objects \\ 
		
		\addlinespace
		
		\makecell{Tang et al. \cite{tang20253d}} & \makecell{2025} & 
		Automates 3D reconstruction from 2D inputs, enhancing 3D scene understanding for dynamic environments & 
		Accuracy drops in sparse or occluded scenes and requires pre-trained VLM with 2D grounding \\ 
		
		\addlinespace
		
		\makecell{Ni et al. \cite{ni2024grid}} & \makecell{2024} & 
		Graph networks model object relations for semantic reasoning, while LLM enables natural instruction decomposition& 
		Graph construction becomes costly in cluttered scenes and limited to pre-defined relationship types \\ 
		
		\addlinespace
		
		\makecell{Ekpo et al. \cite{ekpo2024verigraph}} & \makecell{2024} & 
		Scene graphs enable explicit spatial relationship verification and iterative LLM plan correction ensures physical enforceability  & 
		Graph construction latency and fails on ambiguous spatial queries \\
		
		\addlinespace
		
		\makecell{Mei et al. \cite{mei2024replanvlm}} & \makecell{2024} & 
		Dual-loop correction improves error recovery robustness and maintains plan consistency during dynamic environment changes  & 
		Iterative corrections increase task completion time and dependent on VLM's visual grounding accuracy \\ 
		
		\addlinespace
		
		\makecell{Wu et al. \cite{wu2023embodied}} & \makecell{2023} & 
		Tight LLM-VLM alignment ensures physically feasible plans and context-aware constraint handling  & 
		Requires complete 3D object models for constraint verification and struggles with implicit physical rules \\ 
		
		\addlinespace
		
		\bottomrule
	\end{tabular}
\end{table*}

{
		In spatial representation and semantic comprehension, Huang et al. \cite{huang2023visual} proposed VLMaps, which is a spatial mapping representation that directly combines pre-trained visual language features with 3D reconstruction of the physical world. In this way, service robots can position space targets outside object-centric space targets during mission planning and execution, such as `between the TV and the couch'.
}
 At the same time, Tang et al. \cite{tang20253d} proposed a new framework that uses a 2D prompt synthesis module to enable VLM to train on 2D images and texts, and can independently extract accurate 3D spatial information without manual intervention, thereby significantly enhancing 3D scene understanding, enabling robots to plan and perform actions adaptively in a dynamic environment.

\begin{figure}
	\centering
	\includegraphics[width=.9\columnwidth]{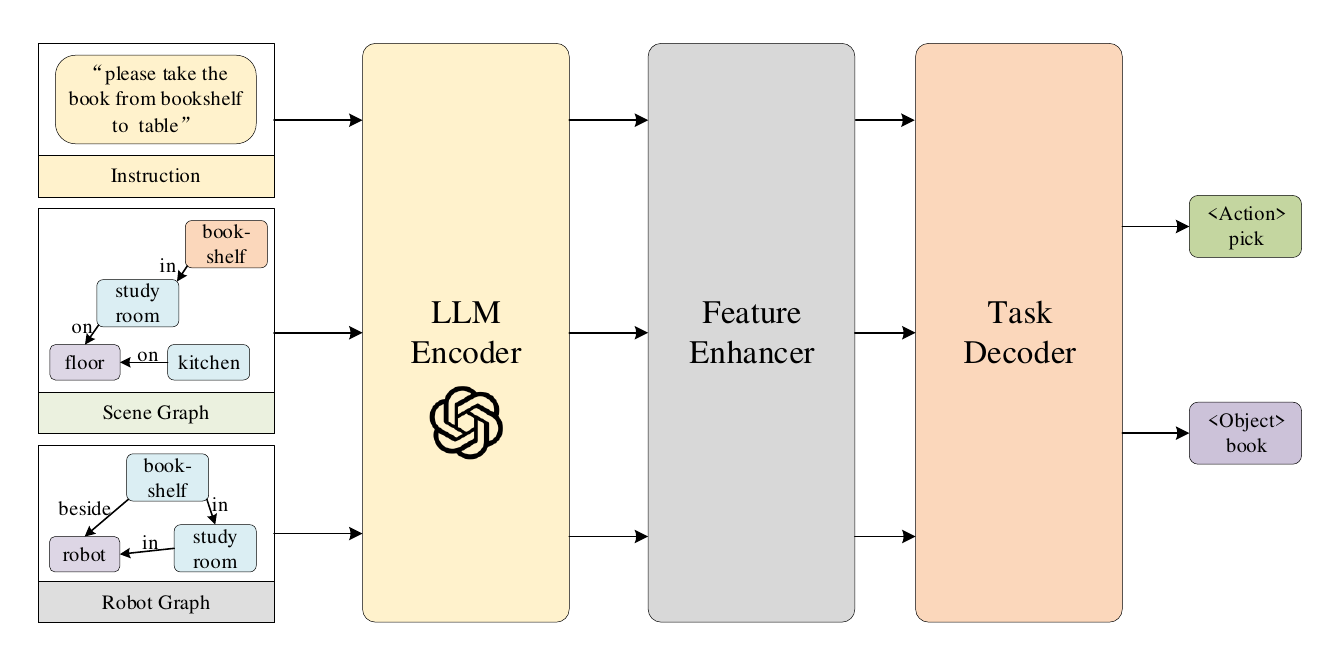}
	\caption{The architecture diagram of GRID.}
	\label{fig:vision1}
\end{figure}

For task decomposition and planning refinement, Ni et al. \cite{ni2024grid} proposed a new method called graph-based robot instruction decomposer (GRID), as shown in Figure \ref{fig:vision1}. This method uses the LLM and graph attention networks to encode the object attributes and relationships in the graph, so that the robot can obtain the semantic knowledge widely observed in the environment from the scene graph. Ekpo et al. \cite{ekpo2024verigraph} proposed VeriGraph, which is a new framework for integrated VLM for robot planning. VeriGraph uses scenario graphs as intermediate representations to capture key objects and spatial relationships, thereby improving planning validation and optimization. The system generates a scene map from the input image and uses it to iteratively check and correct the sequence generated by the LLM-based task planner to ensure compliance with constraints and enforceability.

\begin{figure}
	\centering
	\includegraphics[width=1.0\columnwidth]{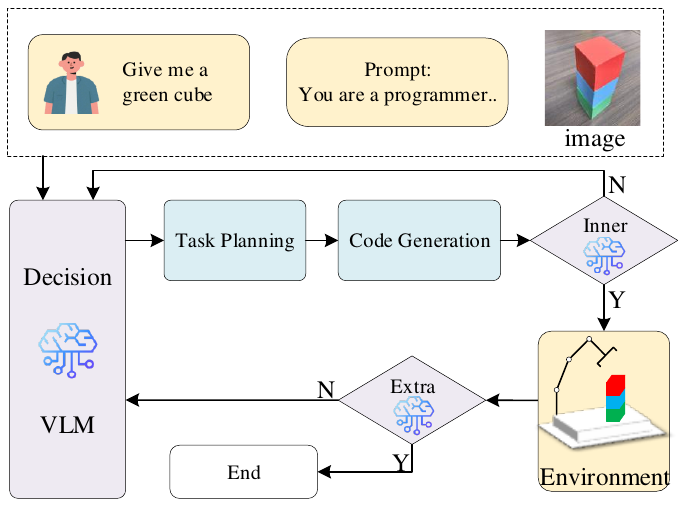}
	\caption{The framework of ReplanVLM.}
	\label{fig:vision2}
\end{figure}

{
	Regarding dynamic adaptation and self-correction, Mei et al. \cite{mei2024replanvlm} proposed a ReplanVLM framework for robot task planning, as shown in Figure \ref{fig:vision2}. An internal error correction mechanism and an external error correction mechanism are proposed for error correction at the corresponding stage.
}
 Shirai et al. \cite{shirai2024vision} proposed a visual language interpreter ViLaln, which is a new problem description framework generated by the most advanced LLM and visual language model. It can optimize the generated problem description through the error message feedback of the symbol planner.

{
		In embodied task execution and physical constraint management, Wu et al. \cite{wu2023embodied}  developed a TaPA for landing planning with physical scene constraints in embodied tasks. The agent generates an executable plan based on the objects in the scene by aligning the LLM with the visual language perception model. Li et al. \cite{li2024fine} proposed the fine-grained task planning FGTP, which aligns the object attributes with their corresponding images through VLM to obtain the actual attributes of the object. This is a method that combines object ontology knowledge with the LLM to create action sequences. 
}
In order to deal with the difficulties encountered in dealing with complex tasks and interacting with the environment effectively, Luan et al. \cite{luan2024enhancing} introduced VLM as an environment-aware sensor and assimilated the results into LLM, thus integrating task objectives with environmental information and enhancing the enforceability of task planning output. Huang et al. \cite{huang2024combining} used the combination of VLM and LLM to perform the task of robot-to-human handover with semantic object knowledge, which enhanced the perception and interaction of robots in a dynamic environment and paved the way for more seamless human-machine collaboration.

Furthermore, Zhang et al. \cite{zhang2025gptarm} developed GPTArm and proposed a robot task processing framework RTPF, which integrates real-time visual perception, situational reasoning and autonomous strategy planning, so that the manipulator can interpret natural language commands, decompose user-defined tasks into executable sub-tasks, and dynamically recover from errors. Wake et al. \cite{wake2024gpt} proposed a ready-to-use multi-mode task planner using VLM and LLM, and introduced a pipeline to enhance the general visual language model GPT-4V to promote the one-time visual teaching of robots, successfully transforming human actions in videos into robot executable programs. Liu et al. \cite{liu2024vision} proposed a VLM-driven method to understand scenes in unknown environments. The visual language model is used as the skeleton of the language model for image description and scene understanding, which further proves that the task planning ability of the service robot in the home environment is significantly improved through the visual language model.

\subsection{Active Task Cognition}

Active task cognition necessitates that robots possess the capability to comprehend their environment and autonomously deduce task-related information. Fundamentally, active task cognition can be characterized as a visual reasoning task, wherein the reasoning process serves as the cornerstone of the cognitive procedure. This form of visual reasoning extends beyond mere object identification within a scene \cite{liu2022behavior}; it encompasses an understanding of the interactions among these objects, as well as their respective significance and function in the real world. This section elaborates on active task cognition from visual reasoning, which is typical as shown in Table \ref{tab3:visual_cognition}.

\begin{table*}[t]
	\centering
	\caption{Typical Literature on Visual Reasoning}
	\label{tab3:visual_cognition}
	\begin{tabular}{>{\centering\arraybackslash}m{2cm} >{\centering\arraybackslash}m{1.5cm} m{5.8cm} m{5.8cm}}
		\toprule
		\multicolumn{1}{c}{\textbf{Paper}} & \multicolumn{1}{c}{\textbf{Year}} & \multicolumn{1}{c}{\textbf{Core Innovation}} & \multicolumn{1}{c}{\textbf{Limitations}}\\
		\midrule
		
		\makecell{Mo et al. \cite{mo2021mutual}} & \makecell{2021} & 
		Proposed a graph-based multimodal framework for joint learning and segmentation & 
		Requires paired multimodal data; sensitive to modality misalignment \\
		
		\addlinespace
		
		\makecell{Liu et al. \cite{liu2021toward}} & \makecell{2021} & 
		Replaces bounding-box features with fine-grained region attention; more precise relationship prediction in scene graphs& 
		Higher computational cost than bounding-box approaches and needs detailed region-level annotations for training \\
		
		\addlinespace
		
		\makecell{Zhang et al. \cite{zhang2022sequential}} & \makecell{2022} & 
		Novel 3D scene graph structure combining geometric and semantic information& 
		Real-time graph updates may be challenging in dynamic environments \\
		
		\addlinespace
		
		\makecell{Xie et al. \cite{xie2022knowledge}} & \makecell{2022} & 
		Integrates visual concepts with external knowledge bases; generates semantically rich questions about images & 
		Relies on manually curated knowledge bases; limited to domains covered by the knowledge base \\
	
		\addlinespace
		
		\makecell{Hao et al. \cite{hao2024embosr}} & \makecell{2024} &
		Chain-of-Thought enhanced spatial reasoning; improves SQA accuracy for complex spatial queries &
		Dependent on LLM reasoning reliability; limited to predefined spatial primitives\\
		
		\addlinespace
		
		\makecell{Cheng et al. \cite{cheng2024spatialrgpt}} & \makecell{2024} & 
		Specialized in geometric relationship reasoning; bridges visual perception with symbolic reasoning & 
		Prone to LLM hallucination in spatial descriptions; requires accurate object detection as input \\
		
		\addlinespace
		
		\bottomrule
	\end{tabular}
\end{table*}

In the realm of visual cognition research, an emerging body of literature underscores the criticality of representing relational structures and semantic information gleaned from images. Mo et al. \cite{mo2021mutual} introduced a graph learning-based multimodal prior guided segmentation framework, which innovatively addresses the dual challenges of feature extraction and cross-modal correspondence. This framework enables the efficient extraction of modality-specific features while establishing robust regional correspondences across multiple modalities, thereby facilitating more comprehensive visual understanding. Yang et al. \cite{yang2021solver} proposed an alternative approach centered on emotional reasoning. Their methodology involves constructing an emotional graph that integrates semantic concepts and visual features. Leveraging Graph Neural Network (GCN) techniques, they perform reasoning operations on the emotional graph, ultimately generating emotion-enhanced object features. These advancements collectively highlight the potential of graph-based models in enriching visual cognition by encoding complex semantic and relational information. 

In addition, integrating the scene graph into the intelligent system not only enhances the visual cognition, but also emphasizes the relationship and attributes of a single object, which promotes the scene understanding \cite{gu2024interactive,mascaro2024scene,senior2025graph}. Liu et al. \cite{liu2021toward} devised a region-aware attention learning approach. This method emphasizes fine-grained visual regions over coarse-grained bounding box features, thus optimizing the image scene graph generation process. Meanwhile, Zhang et al. \cite{zhang2022sequential} innovatively incorporated cooking logic into food image component recognition. By extracting semantic information from images, their method directly generates corresponding recipes and preparation techniques, bridging visual understanding and practical application. Jiao et al. \cite{jiao2022sequential} planned and designed a 3D scene graph representation for the robot sequence by abstracting the scene layout of robot-scene interaction. These models can effectively capture the spatial and semantic relationships of objects to promote the robot's cognition of tasks.

In addition, the application of Visual Question Answering (VQA) technology also contributes to enhancing the extraction of semantic information. Frankli et al. \cite{kenfack2020robotvqa} proposed RobotVQA, an innovative robot vision architecture designed for VQA tasks. The architecture uses RGB or RGBD images of the scene where the robot is located as the original input, and accurately identifies related objects in the scene through advanced object detection algorithms. On this basis, the semantic map of the scene is systematically constructed by analyzing the qualitative spatial relationship, so as to provide structured environmental information support for subsequent intelligent decision-making and task execution. Xie et al. \cite{xie2022knowledge} proposed a knowledge-based VQA model, which innovatively integrates visual concepts and non-visual knowledge to generate deep semantic problems for images, and significantly improves the scene cognition level of robots. Similarly, Hao et al. \cite{hao2024embosr} proposed a novel spatial reasoning paradigm, which significantly improves the performance of Situational Question Answering (SQA) system in spatial relationship reasoning and complex query processing by organically integrating basic model and thinking chain mechanism. This paradigm enables the system to analyze spatial semantic information more accurately through a structured reasoning process. In order to help robots understand scene context information, Wang et al. \cite{wang2024cog} constructed a chain guided learning model for graphical question answering (DQA). The model relies on the LLM to guide the graph analysis tool, which significantly improves the accuracy of graph analysis and enriches the background knowledge system. Coincidentally, Cheng et al. \cite{cheng2024spatialrgpt} proposed spatial region GPT (SpatialRGPT), which is dedicated to spatial perceptual VQA. By strengthening spatial perception and reasoning ability, it can effectively promote the cognitive process of robot tasks. 

{
		In the above research, active task cognition mainly focuses on three technical directions: Graph-based Reasoning Systems: Several studies \cite{mo2021mutual,liu2021toward,zhang2022sequential,jiao2022sequential} have developed sophisticated graph representations that enable robots to actively reason about object relationships and task contexts.  For instance, Jiao et al. \cite{jiao2022sequential} proposed a 3D scene graph representation that abstracts scene layouts and robot-scene interactions, allowing robots to dynamically update their understanding of the environment during task execution.
		VQA Frameworks: Advanced VQA systems \cite{xie2022knowledge,hao2024embosr,kenfack2020robotvqa,wang2024cog} empower robots to actively query their visual understanding.  The RobotVQA system \cite{kenfack2020robotvqa} demonstrates how robots can construct semantic maps by analyzing qualitative spatial relationships through a question-answering paradigm.
		Multimodal Chain-of-Thought Reasoning: New frameworks like SpatialRGPT \cite{cheng2024spatialrgpt} combine visual perception with step-by-step reasoning, enabling robots to actively verify their understanding through multimodal evidence.
		Despite these advances, current systems still face limitations in genuine active cognition.  Most approaches remain reactive rather than proactive, requiring explicit human queries or predefined triggers rather than autonomously identifying task-relevant information. At present, how to construct a reasoning model with active task cognitive ability based on VLM and visual perception information has become a key challenge to be overcome.
}


\section{Audio-based LLM Task Planning}\label{Audio-based LLM Task Planning}

Although visual and textual modalities dominate in robot task planning, audio signals provide a unique channel for the perception of service robots, especially in visual deprivation scenarios (e.g., dark environments or assisted care). Audio-based LLM mission planning leverages voice commands, ambient sound, and acoustic cues to enable robots to interpret implicit user intents and dynamic contexts. However, compared with the vigorous development of LLM task planning based on text and visual input, audio-driven task planning still faces many challenges, and related research is in a stage of rapid development.This part studies how LLMs handle audio input for task planning, and elaborates on the work of Audio Language Model (ALM), acoustic scene perception and speech interaction. Typical examples are shown in Table \ref{tab4:audio_based}.

\begin{figure}
	\centering
	\includegraphics[width=.9\columnwidth]{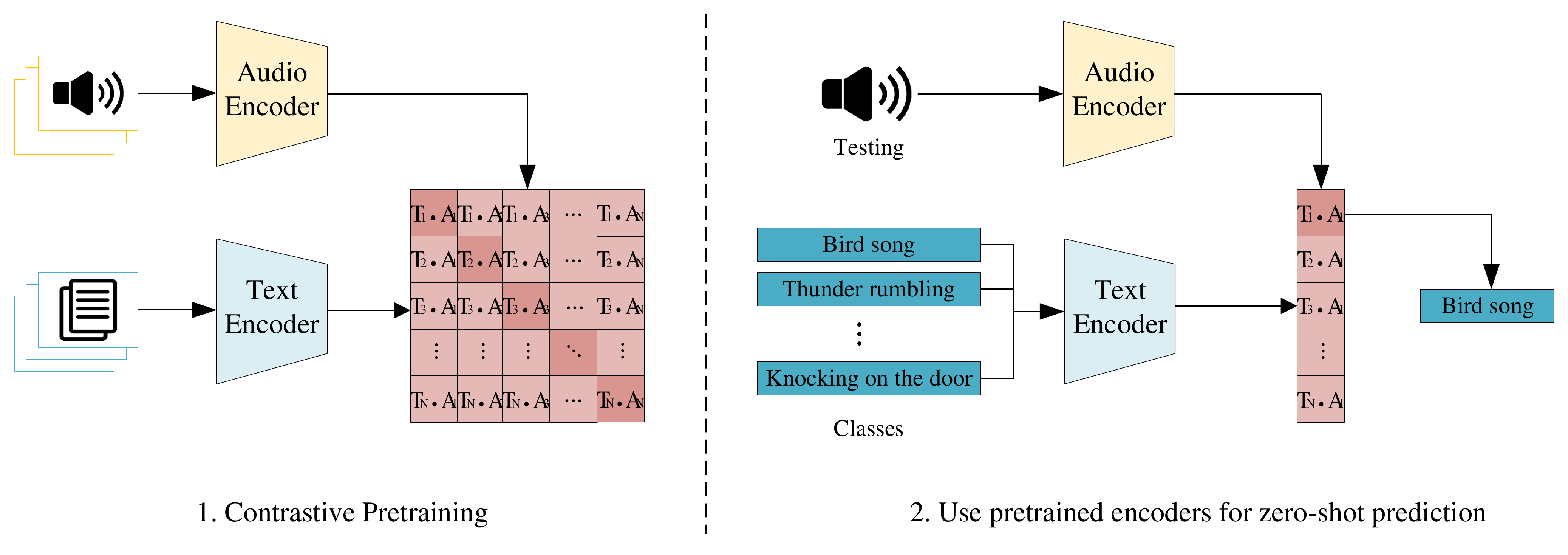}
	\caption{The architecture diagram of CLAP.}
	\label{fig:audio1}
\end{figure}
\begin{figure}
	\centering
	\includegraphics[width=.9\columnwidth]{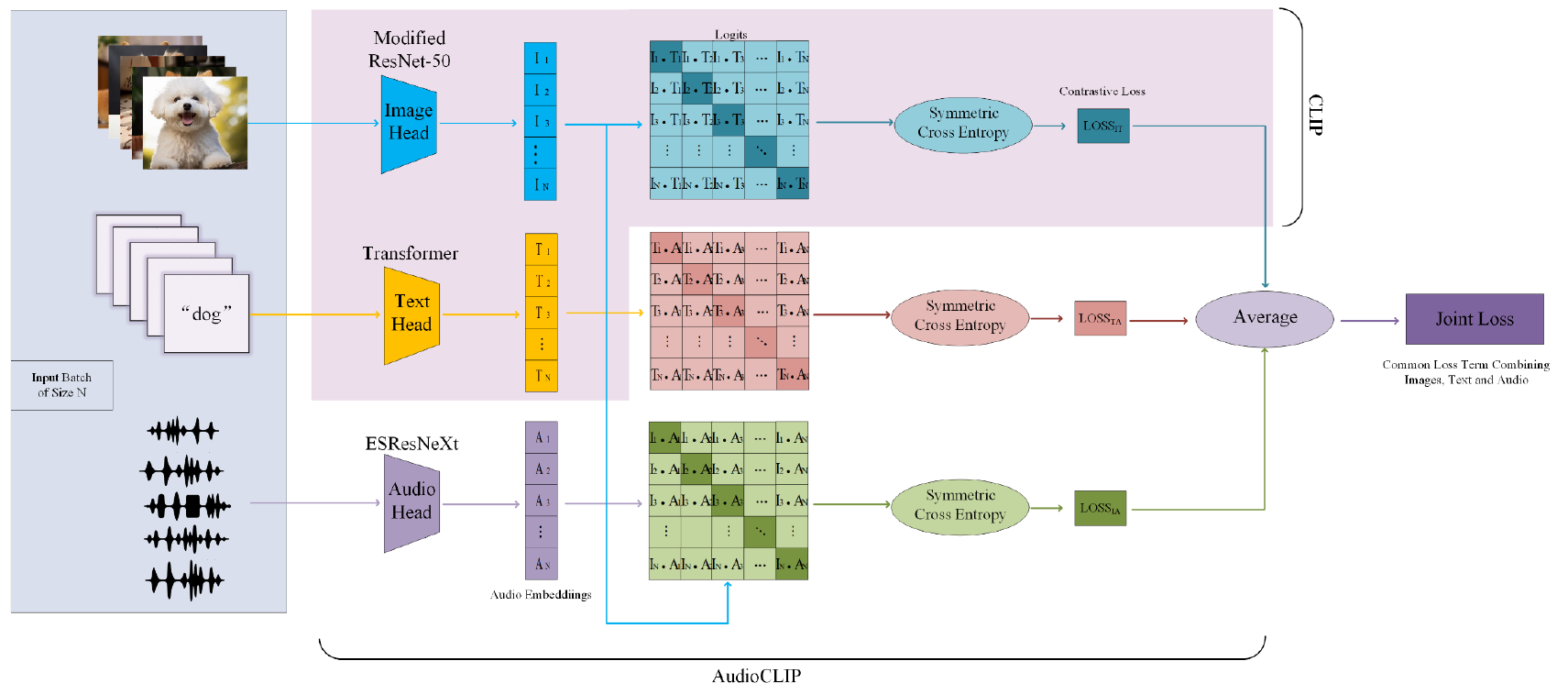}
	\caption{Workflow of the proposed AudioCLIP model.}
	\label{fig:audio2}
\end{figure}
\begin{figure}
	\centering
	\includegraphics[width=.9\columnwidth]{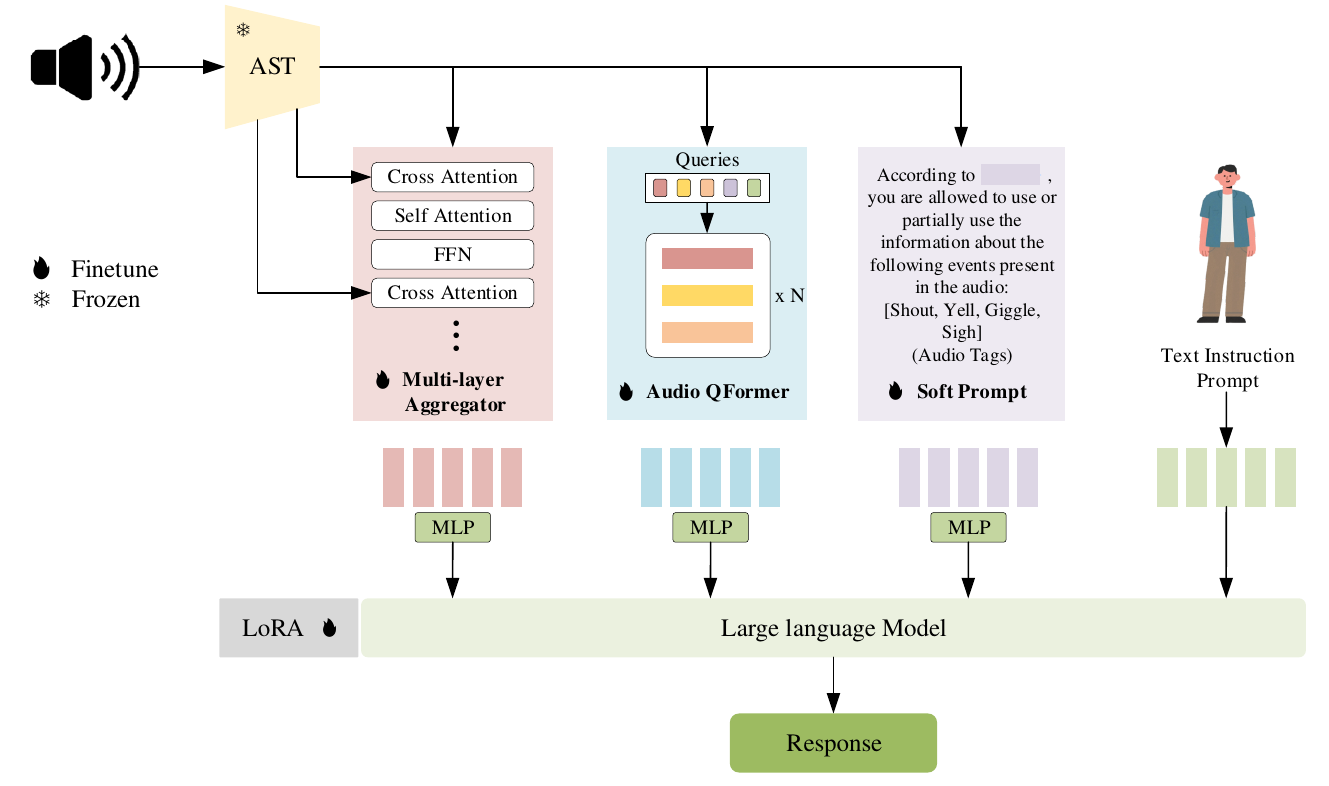}
	\caption{The schematic diagram of GAMA.}
	\label{fig:audio3}
\end{figure}

\begin{table*}[t]
	\centering
	\caption{Typical Literature on Audio-based LLM Task Planning}
	\label{tab4:audio_based}
	\begin{tabular}{>{\centering\arraybackslash}m{2.4cm} >{\centering\arraybackslash}m{1.5cm} m{5.8cm} m{5.8cm}}
		\toprule
		\multicolumn{1}{c}{\textbf{Paper}} & \multicolumn{1}{c}{\textbf{Year}} & \multicolumn{1}{c}{\textbf{Core Innovation}} & \multicolumn{1}{c}{\textbf{Limitations}}\\
		\midrule
		
		\makecell{Elizalde et al. \cite{elizalde2023clap}} & 2023 & 
		Dual-encoder contrastive learning for audio-text alignment; enables zero-shot audio classification and retrieval & 
		Requires large-scale aligned audio-text pairs; struggles with fine-grained audio semantics  \\
		
		\addlinespace
		
		\makecell{Guzhov et al. \cite{guzhov2022audioclip}} & 2022 & 
		Tri-modal fusion via CLIP extension; ESResNeXt audio encoder for environmental sound classification & 
		Limited to coarse sound categories; no joint training of all modalities \\
		
		\addlinespace
		
		\makecell{Kong et al. \cite{kong2024audio}} & 2024 & 
		Few-shot learning for unseen audio tasks; conversational audio reasoning via LLM integration & 
		Context window limits long audio processing; overfits to frequent sound patterns \\
		
		\addlinespace
		
		\makecell{Chen et al. \cite{chen2024salm}} & 2024 & 
		Maintains LLM knowledge while adding audio via adapters; Zero-shot adaptation to new voice instructions & 
		Audio encoder limits real-time processing; degrades with noisy speech \\
		
		\addlinespace
		
		\makecell{Rubenstein et al. \cite{rubenstein2023audiopalm}} & 2023 & 
		Unified PaLM-2 + AudioLM architecture; bidirectional speech-text generation & 
		Relies on high-quality parallel speech-text corpora; generated speech lacks emotional nuance \\
		
		\addlinespace
		
		\makecell{Wang et al. \cite{wang2024can}} & 2024 & 
		Gradient-free adaptation to speakers; in-context learning with few audio-text examples  & 
		Requires representative few-shot examples; limited to vocabulary seen in context \\
		
		\addlinespace
		
		\makecell{Deshmukh et al. \cite{deshmukh2023pengi}} & 2023 & 
		Unified text-generation framework for all audio tasks; eliminates task-specific fine-tuning & 
		No explicit audio-text alignment; struggles with novel sound events \\ 
		
		\addlinespace
		
		\bottomrule
	\end{tabular}
\end{table*}

In terms of the basic model architecture, several breakthrough studies have laid a solid foundation for the joint representation of audio and language. As shown in Figure \ref{fig:audio1}, the CLAP (Contrastive Language-Audio Pretraining) model proposed by Elizalde et al. \cite{elizalde2023clap} effectively connects the language and audio modalities through a dual-encoder structure and a contrastive learning objective, mapping audio and text descriptions into a joint multimodal space to enable flexible class prediction during inference and significantly enhance cross-modal retrieval performance. Building on this work, the team led by Guzhov \cite{guzhov2022audioclip} integrated the high-performance audio model ESResNeXt with the text-image contrastive model CLIP, as shown in Figure \ref{fig:audio2}, constructing a three-modal hybrid architecture that excels in environmental sound classification tasks and extends the zero-shot learning capability of the basic model to the audio domain. Furthermore, Wu et al. \cite{wu2022wav2clip} proposed Wav2CLIP, which innovatively projects audio, images, and text into the shared embedding space of CLIP, offering an efficient solution for tasks like zero-shot sound classification and demonstrating strong adaptability and generalization in multimodal scenarios, thus advancing the field of audio-language joint representation.

To bolster models' comprehension of intricate audio scenes, researchers have engineered diverse specialized architectures. The Audio Flamingo model, devised by Kong et al. \cite{kong2024audio}, exhibits remarkable few-shot learning and conversational prowess, enabling it to rapidly adapt to unseen task scenarios. Gong et al. \cite{gong2023listen} introduced LTU (Listen, Think, and Understand), a novel audio foundation model that demonstrates robust performance and generalization on traditional audio tasks, including classification and captioning. Notably, LTU showcases emergent audio reasoning and understanding capabilities hitherto absent in existing models. As shown in Figure \ref{fig:audio3}, Ghosh et al. \cite{ghosh2024gama} proposed GAMA, a general-purpose large audio language model that integrates advanced audio comprehension and sophisticated reasoning abilities, thereby offering a novel paradigm for the semantic analysis of ambient sounds.

In the realm of voice interaction, numerous studies have focused on enhancing robots' ability to understand and execute spoken instructions. To handle the subtleties of oral commands in scenarios like social navigation, Sun et al. \cite{sun2024beyond} proposed ``Beyond Text'', a method that improves LLM decision-making by integrating audio transcription with functional sub-parts, seamlessly combining text-based guidance with a human-auditory information language model and marking a significant advancement in social robot navigation and broader human-robot interaction. Meanwhile, Chen et al. \cite{chen2024salm} introduced a novel Speech Augmented Language Model (SALM), featuring a frozen text LLM, an audio encoder, a modality adapter module, and LoRA layers to adapt to voice input and task instructions. The unified SALM not only achieves performance comparable to the Conformer baseline for automatic speech recognition (ASR) and speech translation (AST) tasks but also demonstrates zero-shot context learning capabilities, highlighting its adaptability in voice interaction scenarios. Moreover, Rubenstein et al. \cite{rubenstein2023audiopalm} introduced AudioPaLM, a sophisticated large language model designed for speech understanding and generation. AudioPaLM integrates the text-based language model PaLM-2 \cite{anil2023palm} and the speech-based language model AudioLM \cite{borsos2023audiolm} into a unified multimodal architecture. This innovative integration endows AudioPaLM with the ability to process and generate both text and speech, enabling it to perform tasks such as speech recognition and speech-to-speech translation with remarkable efficiency and accuracy. 

To assess the performance of audio comprehension models in tasks demanding expert-level knowledge and intricate reasoning, Sakshi et al. \cite{sakshi2024mmau} introduced MMAU (Multimodal Multilingual Audio Understanding), a benchmark featuring 10,000 meticulously curated samples along with manually annotated natural language questions and answers, covering speech, environmental sounds, and music. In a different yet equally impactful vein, Wang et al. \cite{wang2024can} proposed a novel Speech-based In-context Learning (SICL) approach that harnesses a few context examples—paired spoken words and labels—from specific dialects or speakers to achieve language- or speaker-level model adaptation during testing without gradient descent, effectively resolving the adaptability issues stemming from dialect and speaker disparities. At the application level, Salewski et al. \cite{salewski2023zero} developed ZerAuCap, a model guided by a pre-trained audio language model to generate captions that accurately describe audio content. It utilizes audio context keywords to prompt the language model, enabling the generation of text closely associated with the sounds. Meanwhile, addressing the limitation in generating appropriate language for open-ended tasks like audio captioning and audio Q\&A, Deshmukh et al. \cite{deshmukh2023pengi} introduced Pengi. This innovative model capitalizes on transfer learning by formulating all audio tasks as text generation tasks. Pengi's unified architecture can seamlessly handle both open and closed tasks, eliminating the need for additional fine-tuning or task-specific modifications, thus offering a versatile and efficient solution for various audio-language processing needs. 

Despite these advances, audio LLM planning still faces many challenges. The ClozeGER error correction framework proposed by  Hu et al. \cite{hu2024listen} for speech recognition errors improves the fidelity of the corrected output by introducing a multimodal LLM (SpeechGPT) to receive the source speech as additional input, but its robustness in complex noise environments still needs to be improved.

\section{Multimodal-based LLM Task Planning}\label{Multi-modal Large Language Models Planning}

With the rapid development of LLM, limited by its narrow sensory input and fragile environmental understanding ability, people are no longer satisfied with a single-modal LLM, but try to integrate the input of multiple modalities \cite{ji2025learning} and integrate visual, language, audio and other signals into a MLLM. As a transformative paradigm of robot planning, MLLM enables robots to understand fuzzy instructions, adapt to dynamic environments, and learn from cross-modal correlations. In this section, we will introduce the current research progress and application of multimodal large language models.

\subsection{Multimodal Large Language Models}

\begin{figure}
	\centering
	\includegraphics[width=.9\columnwidth]{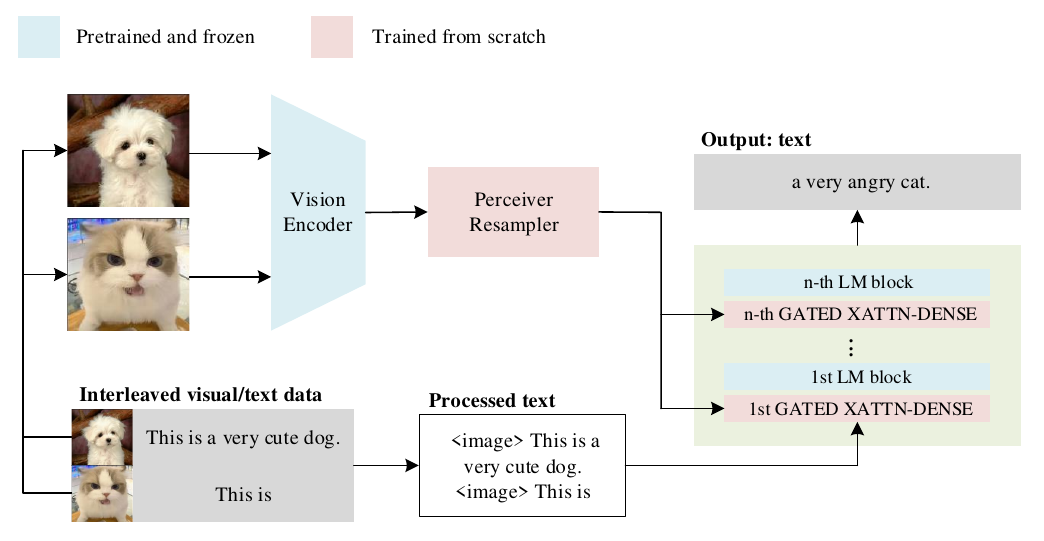}
	\caption{The architecture diagram of Flamingo.}
	\label{fig:mllm1}
\end{figure}

\begin{figure}
	\centering
	\includegraphics[width=.9\columnwidth]{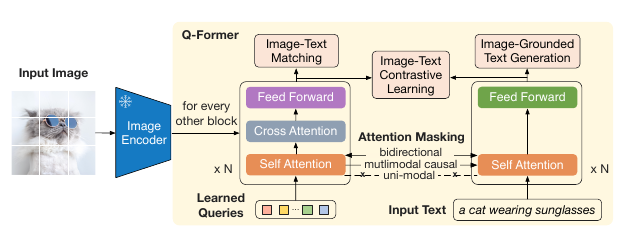}
	\caption{The schematic diagram of BLIP-2. Reproduced with permission from [126]. Copyright 2023, JMLR.org.}
	\label{fig:mllm2}
\end{figure}

In this section, we will explore a series of typical MLLMs and analyze the differences between them. The Flamingo architecture proposed by Alayrac et al. \cite{alayrac2022flamingo} firstly used the query-based cross-attention mechanism, that is, the `sensor resampler'. As shown in Figure \ref{fig:mllm1}, the architecture connects the powerful pre-trained pure visual model with the pure language model, and can handle any alternating visual and text data sequences, as shown in Figure \ref{fig:mllm2}. This innovative way to build a strong visual language interaction module, showing the excellent ability to learn in the context of the environment. Similarly, BLIP-2 \cite{li2023blip} uses a similar method in the image coding process, using the Qformer model to extract image features. Subsequently, the model promotes the interaction between image and text through the cross-attention mechanism. When fine-tuning for a specific downstream task data set, BLIP-2 will unlock the visual encoder and fine-tune it together with Qformer.

In addition, Yang et al. \cite{yang2023mm} proposed MM-REACT, which combines ChatGPT with various visual models to complete multimodal tasks. The system is similar to the previous research results using the visual question answering caption model and the language image model. However, the special feature of MM-REACT is that it has the ability to independently determine whether to call the visual model. The LLaMA-Adapter proposed by Zhang et al. \cite{zhang2023llama} realizes an efficient fine-tuning process through an adapter, thus expanding to multimodal application scenarios. MiniGPT-4 proposed by Zhu et al. \cite{zhu2023minigpt}, combines BLIP-2 and Vicuna, and uses a projection layer to align the frozen visual encoder with the frozen high-level large language model Vicuna, thereby achieving advanced multimodal functions. LLaVA \cite{liu2023visual} uses GPT-4 to generate command fine-tuning data, and further strengthens the command tracking ability of the multimodal large language model by training the visual command adjustment data. LLaVA-1.5 \cite{liu2024improved} is extended on this basis, and the VQA data set is integrated into the instruction adjustment data, which greatly improves the performance in a series of benchmark tests. Kosmos-2, introduced by Peng et al. \cite{peng2023kosmos}, implements new capabilities for perceiving object descriptions (i.e., bounding boxes) and anchoring text to visual scenes by representing reference expressions as links in the Markdown format, where objects are described as sequence of position markers.

In contrast, QWen2-VL \cite{wang2024qwen2} adopts a more complex three-stage training process and introduces a Naive Dynamic Resolution mechanism, so that the model can dynamically process images with different resolutions into different numbers of visual markers. This method enables the model to generate more efficient and accurate visual representation, which is closely related to the human perception process. The mPLUG-Owl introduced by  Ye et al. \cite{ye2023mplug}, through the modular learning of the basic large language model, combined with the visual knowledge module and the visual abstractor module, gives the large language model multimodal capabilities. This method makes full use of the synergy between modalities and improves the performance of plain text tasks and multimodal tasks. At the same time, Otter proposed by Li et al. \cite{li2023mimic} improved the OpenFlamingo model, which focuses on improving the execution of instructions and effectively uses context samples, showing extraordinary proficiency in multimodal perception, reasoning and situational learning. CogCoM, introduced by Qi et al. \cite{qi2024cogcom}, introduces `Chain of Manipulations', which is a mechanism that allows visual language models to solve problems step by step with evidence. After training, the model can solve various visual problems by actively triggering internal operations (such as positioning, magnification) and generating results (such as bounding boxes, images), without the need for external tools, while allowing users to trace the causes of errors. Yan et al. \cite{yan2024vigor} introduced a new framework, Visual Grounding Through Fine-Grained Reward Modeling (ViGoR), which uses fine-grained reward modeling to significantly enhance the visual positioning ability of large-language visual models based on pre-training baselines.

{
		Different modal data have significant differences in terms of presentation, resolution and semantic granularity, and how to unify them to extract common information while retaining their unique details is the core challenge in the fusion process. The above studies have developed a variety of fusion techniques to try to solve these problems, mainly including attention mechanisms, feature alignment, gating mechanisms and graph neural networks.
		Attention mechanisms have emerged as the predominant fusion strategy in Transformer-based architectures. Cross-attention enables dynamic association between modalities \cite{alayrac2022flamingo}, while self-attention captures intra-modal relationships. This dual mechanism allows flexible information interaction tailored to specific task requirements.
		Feature alignment techniques create shared latent spaces for heterogeneous data. Contrastive learning methods like CLIP \cite{radford2021learning} align modalities by maximizing similarity of matched pairs while minimizing unrelated ones. The Q-Former module in BLIP-2 \cite{li2023blip} extends this through shared query vectors that learn unified cross-modal representations.
		Gating mechanisms provide dynamic information flow control. Lightweight adapters, as seen in LLaMA-Adapter \cite{zhang2023llama}, enable context-aware feature injection without full model fine-tuning. These systems automatically adjust modality weights based on environmental conditions.
		Graph Neural Networks offer structural fusion capabilities by representing multimodal elements as graph nodes with spatial/semantic edges. Systems like GRID \cite{ni2024grid} and VeriGraph \cite{ekpo2024verigraph} employ Graph Attention Networks to combine visual scene graphs with verbal commands, significantly enhancing spatial reasoning for robotic tasks.
}

\subsection{Technological Evolution and Practical Applications}

\begin{figure}
	\centering
	\includegraphics[width=.9\columnwidth]{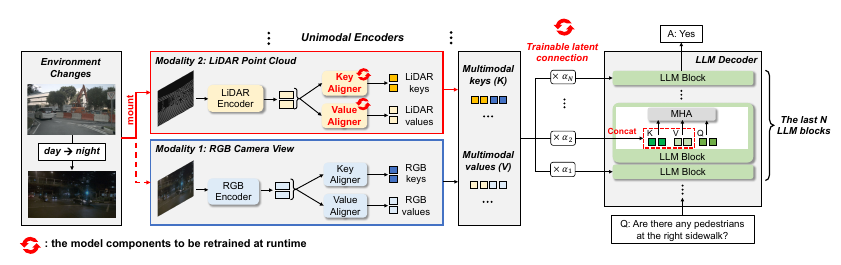}
	\caption{Example of modal adaptation of mPnP-LLM to two input modes. Reproduced with permission from [139]. Copyright 2023, licensed under CC BY 4.0, ACM.}
	\label{fig:mllm3}
\end{figure}
\begin{figure}
	\centering
	\includegraphics[width=.9\columnwidth]{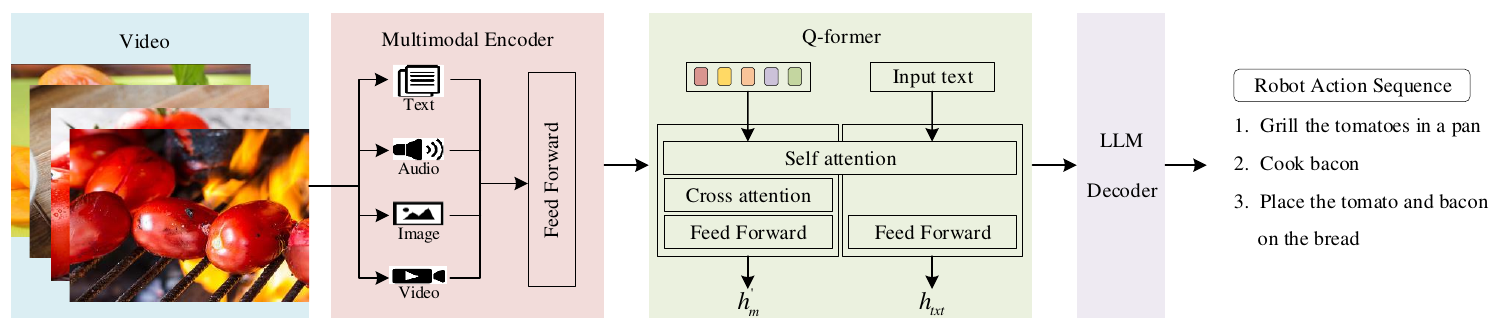}
	\caption{The schematic diagram of AVBLIP.}
	\label{fig:mllm4}
\end{figure}
With the continuous evolution of artificial intelligence models, service robots have become a key area for the application of artificial intelligence technology. With its excellent perception and reasoning efficiency, MLLMs, as the core intelligent center of service robots, play an indispensable role in action command generation and task planning. This section will discuss in depth from multiple dimensions such as model evaluation optimization, robot system integration, and task planning interaction, and systematically analyze the innovative practices and technological breakthroughs of MLLMs in service robot task planning scenarios.

In the field of model evaluation and optimization, researchers are committed to improving the performance and efficiency of MLLMs. The OmniBench benchmark proposed by Li et al. \cite{li2024omnibench} aims to comprehensively evaluate the parallel processing and reasoning ability of the model for multimodal inputs such as visual, auditory, and text. By constructing a standardized evaluation system, it provides an important basis for quantifying the performance of the model. Aiming at the problem of insufficient multimodal processing efficiency in resource-constrained scenarios, the mPnP-LLM model developed by Huang et al. \cite{huang2023modality} creatively combines single-modal encoders with dynamically configurable LLM modules and supports dynamic training at runtime, as shown in Figure \ref{fig:mllm3}. This design enables the model to independently select effective modes according to actual needs, which greatly improves resource utilization efficiency. In the face of the application bottleneck caused by the model 's dependence on specific input formats, the LLMBind framework proposed by Zhu et al. \cite{zhu2024llmbind} transforms multimodal input into task-specific token sequences with the help of expert mixing (MoE) mechanism, and realizes efficient processing and output conversion of multimodal data such as image, text, video and audio. In addition, the RespLLM framework proposed by Zhang et al. \cite{zhang2024respllm}, another way to unify text and audio representations, is applied to respiratory health prediction tasks. The framework makes full use of the rich prior knowledge of pre-trained LLM, and realizes the deep fusion of audio and text information through cross-modal attention mechanism, which provides a new technical paradigm for multimodal model to process heterogeneous data.

At the application level of robot systems, many studies have focused on the deep integration of MLLMs into service robot systems. The service robot system constructed by  Ni et al. \cite{ni2024design} integrates hardware design, scene modeling algorithm, mobile navigation strategy based on TEB path planning, grasping operation based on pre-training model, and task planning module based on visual language model, forming a complete technical chain and realizing the whole process intelligence from perception to decision-making. The multimodal language model agent system proposed by Chung et al. \cite{chung2024empowering} combines the task planning function with the interaction ability of the physical robot by integrating the router, the chat robot and the task planner, which significantly improves the decision-making efficiency and interaction flexibility of the robot in practical applications. The research work of  Jiang et al. \cite{jiang2023vima} proposes an innovative multimodal prompt formula, which transforms diversified robot operation tasks into a unified sequence modeling problem, and establishes an evaluation protocol including multimodal tasks and system generalization ability, which provides a new direction for the standardization research of robot task planning.

In terms of task planning and environmental interaction, researchers continue to explore how to make robots better adapt to complex environments. The Inner Monologue mechanism proposed by Huang et al. \cite{huang2022inner}, by integrating multimodal environmental feedback information, can generate a task planning scheme that is more suitable for the actual scene, effectively coordinate the input information of different sensors, and enhance the AI's ability to understand and respond to the environment. The system developed by Liu et al. \cite{liu2024enhancing} gives full play to the advantages of text understanding and visual processing of MLLMs, and can accurately transform user dialogue and environmental visual information into a robot executable operation plan. The framework proposed by Wang et al. \cite{wang2024large}, with the help of multimodal GPT-4V, deeply integrates natural language instructions with robot visual perception, which significantly improves the accuracy and flexibility of embodied mission planning. The PaLM-E embodied language model proposed by Driess et al. \cite{driess2023palm} realizes the direct connection between sensor data and language model in the real world through staggered vision, continuous state estimation and text input coding, and builds a bridge between perceptual information and semantic understanding. As shown in Figure \ref{fig:mllm4}, the AVBLIP model proposed by Kambara et al. \cite{kambara2024human}, by introducing a multimodal encoder, supports joint input of video, audio, voice and text, and can efficiently generate robot action sequences. The multimodal interaction framework proposed by Lai et al. \cite{lai2025nmm}, integrates voice commands and posture information, combines the global information of the environment obtained by the visual system, uses a large-scale language model to analyze the speech text and bounding box data, and avoids the model to generate unreasonable output through key control syntax constraints, thereby generating reliable robot action sequences. In addition, the service robot task reasoning mechanism proposed by Tian et al. \cite{tian11task}, innovatively integrates multimodal information and ontology knowledge, systematically manages user and environment information through ontology knowledge base, collects and integrates multi-modal data such as vision, speech and scene knowledge in real time, and conducts deep reasoning based on fine-tuning LLM, which provides richer and more reliable knowledge support for robot decision-making.

{
	In terms of closed-loop integration, systems like Inner Monologue \cite{chung2024empowering} demonstrate how continuous visual feedback from RGB-D sensors can be integrated with visual-language models to dynamically update task plans when environmental changes are detected. For human-robot collaboration scenarios, approaches such as AudioPaLM \cite{rubenstein2023audiopalm} enable natural language corrections during task execution, allowing users to verbally adjust the robot's actions when deviations occur. The integration of reinforcement learning from human feedback (RLHF \cite{kirk2023understanding}) provides another important mechanism for adapting LLM outputs to real-world uncertainties, where human demonstrations and corrections help align the model's planning with physical constraints and task requirements. These developments highlight the critical need to move beyond static, open-loop task specifications and instead develop systems that can maintain plan feasibility through continuous perception-action cycles and human interaction.
}	

{
		Furthermore, the recent breakthroughs in embodied intelligence, exemplified by Google's Gemini Robotics \cite{team2025gemini} and NVIDIA's GR00T \cite{bjorck2025gr00t}, represent a paradigm shift from traditional LLMs by fundamentally bridging the gap between cognitive AI and physical robotic execution. Unlike conventional LLMs, which are confined to textual data and generate abstract outputs like language or code without direct interaction with the physical world, these novel models are designed as Vision-Language-Action (VLA) systems that integrate perception, understanding, and action into a cohesive whole. Google's Gemini Robotics \cite{team2025gemini}, built upon the Gemini 2.0 \cite{team2024gemini} model, operates as an on-device VLA architecture, processing real-time sensor inputs such as RGB-D and LiDAR to directly output executable joint control commands, thereby bypassing intermediate symbolic planning and significantly reducing latency for enhanced real-time performance and privacy. Concurrently, NVIDIA's GR00T \cite{bjorck2025gr00t} employs a dual-system architecture where a vision-language module interprets the environment before a diffusion transformer generates fluid motor actions, further distinguished by its use of synthetic data generation through GR00T-Dreams, which compresses skill acquisition from months to 36 hours, and its seamless Sim2Real transfer via Isaac Sim 5.0. This architectural and methodological innovation allows both systems to transcend the limitations of traditional cloud-dependent LLMs by embedding physical feasibility checks and uncertainty-aware planning directly into their decision-making processes. The result is an embodied intelligence capable of sensing the physical environment, interpreting high-level task intent from natural language, and translating it into safe, executable physical actions-a critical advancement for applications ranging from industrial automation to service robotics in unstructured environments, thereby tightly coupling the cognitive prowess of AI with the physical dexterity of robotic systems.
}	
\begin{figure*}[t!]
	\centering
    \includegraphics[width=.9\textwidth]{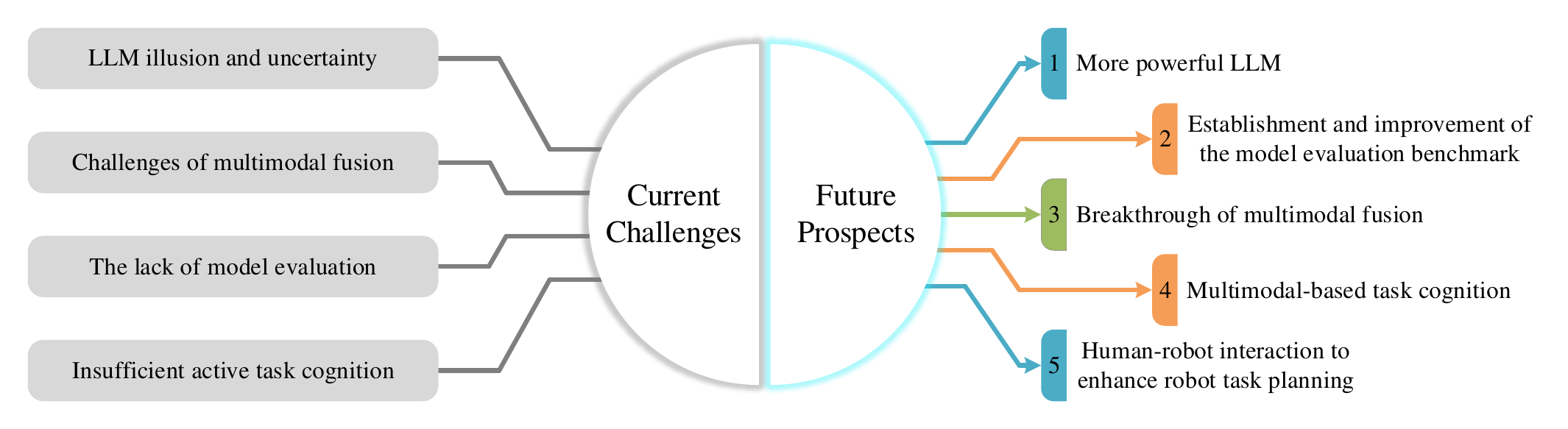}
	\caption{Some current challenges and future perspectives in task planning for service robots.}
	\label{fig:challenges_and_prospects}
\end{figure*}

\section{Current Challenges and Future Prospects}\label{Current Challenges and Future Prospects}

From the review and analysis, it is obvious that significant efforts have been made to improve the planning ability of service robots, and some notable progress has been achieved. However, due to the complexity and diversity of the domestic environment, task planning for service robots still requires substantial exploration. As shown in the Figure \ref{fig:challenges_and_prospects}, the main problems in the field of task planning of service robots based on LLM can be summarized as follows. 

\begin{itemize}
	\item 
	{Insufficiency in real-time performance and safety: The inference processes of current LLMs often demand considerable computational power and time, thereby limiting the ability to guarantee real-time decision-making. How to realize efficient computation of LLM is a general step towards real-time planning. Compounding this, the inherent latency, coupled with the potential for hallucinations and uncertainties during the planning process, poses increased safety risks for robots during task execution. }
	
	\item Challenges of multimodal fusion: It is still a technical problem to effectively fuse text, vision, speech and other modal information and establish a reasonable fusion strategy. How to design an efficient fusion architecture and balance the weights between different modalities needs further study. 
	
	\item Model evaluation and optimization: The lack of evaluation criteria and tools for multimodal LLM makes it difficult to comprehensively evaluate the performance and robustness of the model. At the same time, how to optimize the model structure and improve the computational efficiency and generalization ability is also a problem to be solved.
	
	\item Insufficient active task cognition: At present, LLM still has shortcomings in active task cognition. The service robot can only perform task planning and execution according to the user's instructions, and it is difficult to perform autonomous reasoning and planning according to environmental information and task objectives. How to construct a reasoning model with active task cognitive ability needs further exploration.
	
	\item
	{Tactile language models are underdeveloped: Compared to LLMs, VLMs, and MLLMs, tactile language models remain in their early stages of development. Challenges include scarce tactile datasets, underdeveloped multimodal fusion architectures, and the need for breakthroughs in aligning dynamic contact state representations with language understanding.}
\end{itemize}

In view of the existing problems, the proposed research directions of LLM-based task planning for domestic service robots include: 
\begin{itemize}
\item {More powerful LLM for robot task planning: With the continuous development of technology, the performance of LLM has significantly improved \cite{achiam2023gpt,team2024gemini,grattafiori2024llama}. 
In order to enable LLM to respond in real-time \cite{zheng2025llm,sun2025llm,chen2025robogpt}, it is necessary to explore various strategies such as model compression and acceleration techniques \cite{saha2025vision,sarridis2025indistill}, edge computing deployment \cite{liu2025enhancing}, and more efficient prompt engineering \cite{brown2020language}.
Considering the security of task planning, recent studies have also made substantial progress in mitigating hallucinations in LLMs \cite{friel2023chainpoll,martino2023knowledge,wei2024measuring}. It is also extremely essential to ensure that the generated plans comply with physical and safety constraints \cite{silver2024generalized,wei2022chain,li2024fine}. To further enhance their generalization ability and robustness in robot task planning and executing service-oriented tasks, it remains essential to incorporate external knowledge bases and complementary techniques \cite{ding2023integrating,hanheide2017robot,jiang2019open}.}
     
\item  Establishment and improvement of the model evaluation benchmark: LLMs have been widely used in the field of robotics. Although numerous evaluation benchmarks exist to assess their perception and planning capabilities \cite{zhou2023don,white2024livebench,wang2024benchmark,wang2023pandalm}, there is a lack of a unified and recognized benchmark for single-modal and multimodel, so that each model data can be quantified. Therefore, it is necessary to establish specific evaluation benchmarks tailored to the task planning field of service robots, such as task success rate, planning efficiency, robustness, etc. This would facilitate a more accurate and comparative assessment of model performance. 
   
\item  Breakthrough of multimodal fusion: The text-based LLM has limitations in spatial perception and dynamic adaptation \cite{liu2023llm+}, while visual-based and auditory-based models have difficulties in understanding abstract concepts and extracting effective information in noisy environments, respectively \cite{sun2024beyond,shirai2024vision}. Therefore, the effective fusion of text, visual, auditory and other modal information and the establishment of a reasonable fusion strategy are the key to improving the task planning ability of service robots \cite{alayrac2022flamingo,li2023blip,zhang2024respllm}. For instance, exploring more efficient modal alignment methods and cross-modal attention mechanisms could enable effective interaction and fusion across different types of modal information \cite{ding2025decoupling,han2025multimodal}. 

\item  Multimodal-based active task cognition: Numerous studies have demonstrated the feasibility of integrating LLM with task cognition \cite{chen2022lako,li2023blip,xiao2024florence}. Through the combination of vision, auditory and LLM, service robots can gain stronger active task cognitive ability, enabling them to make autonomous reasoning and planning according to environmental information, sound signals and task objectives, while adapting to the dynamic changing environments. To this end, further research can be conducted on how to deeply integrate multimodal information with task cognition, such as by developing a unified multimodal task cognition model or incorporating task cognition into the pre-training process of multimodal LLMs.

\item  Human-robot interaction to enhance robot task planning: Human-robot interaction is another critical avenue for enhancing the task planning capabilities of service robots, alongside multimodal information. Beyond perceiving the environment through vision and audio, service robots can infer user intentions through direct dialogue and interaction, enabling task planning that better aligns with human preferences \cite{lv2022deep,zhen2023human,duan2024human}. By integrating technologies such as speech recognition, natural language processing, and affective computing, more natural and efficient human-robot interactions can be achieved, thereby significantly improving the effectiveness of task planning.

\item {Advancing tactile intelligence in robotic systems: Future research should focus on three key areas: developing comprehensive tactile-linguistic datasets \cite{yang2024binding}, creating efficient architectures for real-time multimodal fusion \cite{hao2025tla}, and advancing closed-loop tactile control in embodied LLMs\cite{ma2025cltp}. These advancements will enable more adaptive robotic interaction through enhanced multimodal understanding and responsive tactile integration, particularly in precision tasks requiring delicate manipulation.}
\end{itemize}

\section{Conclusion}\label{Conclusion}
\begin{figure*}[t!]
	\centering
	\includegraphics[width=.9\textwidth]{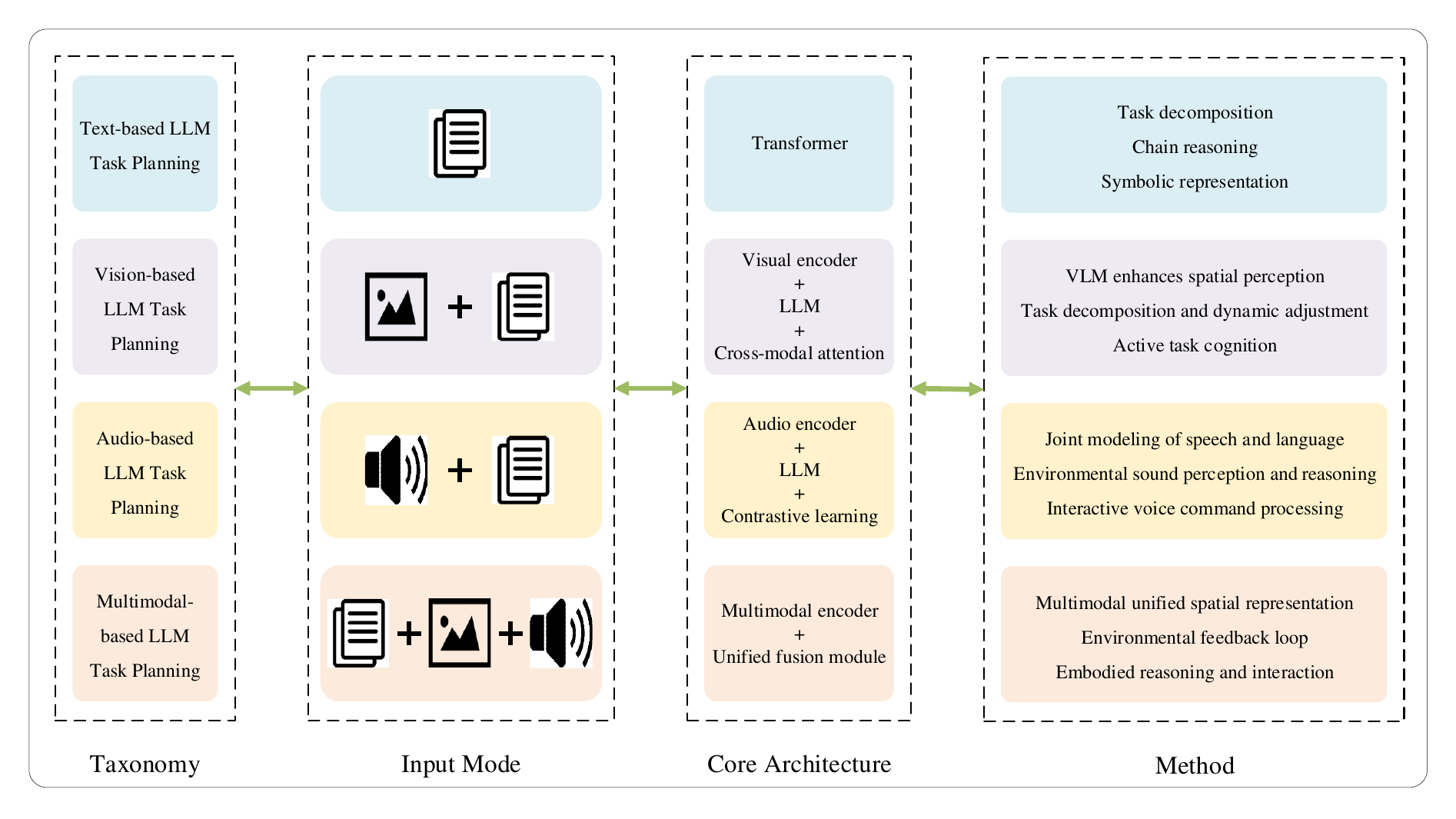}
	\caption{Task planning for different input modalities.}
	\label{fig:conclusion}
\end{figure*}

This paper systematically reviews the current task planning methods for service robots based on LLMs, with a particular focus on key challenges and recent advancements. Specifically, the basic knowledge of LLM is introduced. Then, from the unique perspective of input modality difference, the research trends of LLMs based on text, vision, audio and multimodal fusion in task planning are discussed in depth, as shown in Figure \ref{fig:conclusion}.
Also, a comprehensive and detailed analysis of the existing systems, frameworks and models is carried out, as well as the bottlenecks and problems faced by current research. Finally, constructive and forward-looking insights are put forward on the future research direction of this field. It is anticipated that the findings of this paper can support the development of more robust, reliable, adaptable, and efficient LLM-based task planning methods for service robots, thereby promoting the field to a new level.

\section*{CRediT authorship contribution statement}
\textbf{Shaohan Bian}: Writing – review \& editing, Writing – original draft, Investigation, Formal analysis. \textbf{Ying Zhang}: Writing – review \& editing, Writing – original draft, Investigation, Supervision, . \textbf{Guohui Tian}: Writing – review \& editing, Investigation. \textbf{Zhiqiang Miao}: Writing – review \& editing, Writing – original draft, Investigation. \textbf{Edmond Q. Wu}: Writing – review \& editing, Investigation. \textbf{Simon X. Yang}: Writing – review \& editing, Investigation. \textbf{Changchun Hua}: Writing – review \& editing, Investigation, Supervision.

\section*{Declaration of competing interest}
The authors declare that they have no known competing financial interests or personal relationships that could have appeared to influence the work reported in this paper.

\section*{Acknowledgments}
This work was supported in part by the National Natural Science Foundation of China (62203378, and 62203377), in part by the Hebei Natural Science Foundation (F2024203036, and F2024203115), in part by the Science Research Project of Hebei Education Department (BJK2024195), and in part by the S\&T Program of Hebei (236Z2002G, and 236Z1603G).


\bibliographystyle{unsrt}
\bibliography{cas-refs}

\begin{thebibliography}{100}

\bibitem{bauer2024challenges}
Dominik Bauer, Peter H{\"o}nig, Jean-Baptiste Weibel, Jos{\'e}
  Garc{\'\i}a-Rodr{\'\i}guez, Markus Vincze, et~al.
\newblock Challenges for monocular 6d object pose estimation in robotics.
\newblock {\em IEEE Transactions on Robotics}, 2024.

\bibitem{zhang2025environment}
Ying Zhang, Guohui Tian, Cui-Hua Zhang, Changchun Hua, Weili Ding, and Choon~Ki
  Ahn.
\newblock Environment modeling for service robots from a task execution
  perspective.
\newblock {\em IEEE/CAA Journal of Automatica Sinica}, 2025.

\bibitem{dafarra2024icub3}
Stefano Dafarra, Ugo Pattacini, Giulio Romualdi, Lorenzo Rapetti, Riccardo
  Grieco, Kourosh Darvish, Gianluca Milani, Enrico Valli, Ines Sorrentino,
  Paolo~Maria Viceconte, et~al.
\newblock icub3 avatar system: Enabling remote fully immersive embodiment of
  humanoid robots.
\newblock {\em Science Robotics}, 9(86):eadh3834, 2024.

\bibitem{newbury2023deep}
Rhys Newbury, Morris Gu, Lachlan Chumbley, Arsalan Mousavian, Clemens Eppner,
  J{\"u}rgen Leitner, Jeannette Bohg, Antonio Morales, Tamim Asfour, Danica
  Kragic, et~al.
\newblock Deep learning approaches to grasp synthesis: A review.
\newblock {\em IEEE Transactions on Robotics}, 39(5):3994--4015, 2023.

\bibitem{xu2024grasp}
Kechun Xu, Zhongxiang Zhou, Jun Wu, Haojian Lu, Rong Xiong, and Yue Wang.
\newblock Grasp, see and place: Efficient unknown object rearrangement with
  policy structure prior.
\newblock {\em IEEE Transactions on Robotics}, 2024.

\bibitem{zhang2022semantic}
Ying Zhang, Guohui Tian, Xuyang Shao, Mengyang Zhang, and Shaopeng Liu.
\newblock Semantic grounding for long-term autonomy of mobile robots toward
  dynamic object search in home environments.
\newblock {\em IEEE Transactions on Industrial Electronics}, 70(2):1655--1665,
  2022.

\bibitem{natarajan2024trust}
Manisha Natarajan and Matthew Gombolay.
\newblock Trust and dependence on robotic decision support.
\newblock {\em IEEE Transactions on Robotics}, 2024.

\bibitem{li2025bathing}
Jian Li, Yadong Mo, Shijie Jiang, Lifang Ma, Ying Zhang, and Shimin Wei.
\newblock Bathing assistive devices and robots for the elderly.
\newblock {\em Biomimetic Intelligence and Robotics}, page 100218, 2025.

\bibitem{zhang2020exploring}
Ying Zhang, Guohui Tian, and Huanzhao Chen.
\newblock Exploring the cognitive process for service task in smart home: A
  robot service mechanism.
\newblock {\em Future Generation Computer Systems}, 102:588--602, 2020.

\bibitem{wang2022generalizable}
Chen Wang, Danfei Xu, and Li~Fei-Fei.
\newblock Generalizable task planning through representation pretraining.
\newblock {\em IEEE Robotics and Automation Letters}, 7(3):8299--8306, 2022.

\bibitem{zheng2024knowledge}
Deshuai Zheng, Jin Yan, Tao Xue, and Yong Liu.
\newblock A knowledge-based task planning approach for robot multi-task
  manipulation.
\newblock {\em Complex \& Intelligent Systems}, 10(1):193--206, 2024.

\bibitem{odense2022neural}
Simon Odense, Kamal Gupta, and William~G Macready.
\newblock Neural-guided runtime prediction of planners for improved motion and
  task planning with graph neural networks.
\newblock In {\em 2022 IEEE/RSJ International Conference on Intelligent Robots
  and Systems (IROS)}, pages 12471--12478. IEEE, 2022.

\bibitem{haslum2019introduction}
Patrik Haslum, Nir Lipovetzky, Daniele Magazzeni, Christian Muise, Ronald
  Brachman, Francesca Rossi, and Peter Stone.
\newblock {\em An introduction to the planning domain definition language},
  volume~13.
\newblock Springer, 2019.

\bibitem{zhang2022building}
Ying Zhang, Guohui Tian, Xuyang Shao, Shaopeng Liu, Mengyang Zhang, and Peng
  Duan.
\newblock Building metric-topological map to efficient object search for mobile
  robot.
\newblock {\em IEEE Transactions on Industrial Electronics}, 69(7):7076--7087,
  2022.

\bibitem{zhao2021hierarchical}
Wenrui Zhao and Weidong Chen.
\newblock Hierarchical pomdp planning for object manipulation in clutter.
\newblock {\em Robotics and Autonomous Systems}, 139:103736, 2021.

\bibitem{liu2022service}
Shaopeng Liu, Guohui Tian, Ying Zhang, Mengyang Zhang, and Shuo Liu.
\newblock Service planning oriented efficient object search: A knowledge-based
  framework for home service robot.
\newblock {\em Expert Systems with Applications}, 187:115853, 2022.

\bibitem{zhang2019efficient}
Ying Zhang, Guohui Tian, Jiaxing Lu, Mengyang Zhang, and Senyan Zhang.
\newblock Efficient dynamic object search in home environment by mobile robot:
  A priori knowledge-based approach.
\newblock {\em IEEE Transactions on Vehicular Technology}, 68(10):9466--9477,
  2019.

\bibitem{sun2024leveraging}
Shilong Sun, Chiyao Li, Zida Zhao, Haodong Huang, and Wenfu Xu.
\newblock Leveraging large language models for comprehensive locomotion control
  in humanoid robots design.
\newblock {\em Biomimetic Intelligence and Robotics}, 4(4):100187, 2024.

\bibitem{kim2024survey}
Yeseung Kim, Dohyun Kim, Jieun Choi, Jisang Park, Nayoung Oh, and Daehyung
  Park.
\newblock A survey on integration of large language models with intelligent
  robots.
\newblock {\em Intelligent Service Robotics}, 17(5):1091--1107, 2024.

\bibitem{zeng2023large}
Fanlong Zeng, Wensheng Gan, Yongheng Wang, Ning Liu, and Philip~S Yu.
\newblock Large language models for robotics: A survey.
\newblock {\em arXiv preprint arXiv:2311.07226}, 2023.

\bibitem{li2025large}
Peihan Li, Zijian An, Shams Abrar, and Lifeng Zhou.
\newblock Large language models for multi-robot systems: A survey.
\newblock {\em arXiv preprint arXiv:2502.03814}, 2025.

\bibitem{cui2025task}
Yongcheng Cui, Ying Zhang, Cui-Hua Zhang, and Simon~X Yang.
\newblock Task cognition and planning for service robots.
\newblock {\em Intelligence \& Robotics}, 5(1):119--142, 2025.

\bibitem{mavrogiannis2023core}
Christoforos Mavrogiannis, Francesca Baldini, Allan Wang, Dapeng Zhao, Pete
  Trautman, Aaron Steinfeld, and Jean Oh.
\newblock Core challenges of social robot navigation: A survey.
\newblock {\em ACM Transactions on Human-Robot Interaction}, 12(3):1--39, 2023.

\bibitem{loganathan2023systematic}
Anbalagan Loganathan and Nur~Syazreen Ahmad.
\newblock A systematic review on recent advances in autonomous mobile robot
  navigation.
\newblock {\em Engineering Science and Technology, an International Journal},
  40:101343, 2023.

\bibitem{zhang2023large}
Ceng Zhang, Junxin Chen, Jiatong Li, Yanhong Peng, and Zebing Mao.
\newblock Large language models for human--robot interaction: A review.
\newblock {\em Biomimetic Intelligence and Robotics}, 3(4):100131, 2023.

\bibitem{yang2023foundation}
Sherry Yang, Ofir Nachum, Yilun Du, Jason Wei, Pieter Abbeel, and Dale
  Schuurmans.
\newblock Foundation models for decision making: Problems, methods, and
  opportunities.
\newblock {\em arXiv preprint arXiv:2303.04129}, 2023.

\bibitem{sun2023survey}
Jiankai Sun, Chuanyang Zheng, Enze Xie, Zhengying Liu, Ruihang Chu, Jianing
  Qiu, Jiaqi Xu, Mingyu Ding, Hongyang Li, Mengzhe Geng, et~al.
\newblock A survey of reasoning with foundation models.
\newblock {\em arXiv preprint arXiv:2312.11562}, 2023.

\bibitem{qin2024tool}
Yujia Qin, Shengding Hu, Yankai Lin, Weize Chen, Ning Ding, Ganqu Cui, Zheni
  Zeng, Xuanhe Zhou, Yufei Huang, Chaojun Xiao, et~al.
\newblock Tool learning with foundation models.
\newblock {\em ACM Computing Surveys}, 57(4):1--40, 2024.

\bibitem{vaswani2017attention}
Ashish Vaswani, Noam Shazeer, Niki Parmar, Jakob Uszkoreit, Llion Jones,
  Aidan~N Gomez, {\L}ukasz Kaiser, and Illia Polosukhin.
\newblock Attention is all you need.
\newblock {\em Advances in neural information processing systems}, 30, 2017.

\bibitem{radford2018improving}
Alec Radford, Karthik Narasimhan, Tim Salimans, Ilya Sutskever, et~al.
\newblock Improving language understanding by generative pre-training.
\newblock 2018.

\bibitem{devlin2019bert}
Jacob Devlin, Ming-Wei Chang, Kenton Lee, and Kristina Toutanova.
\newblock Bert: Pre-training of deep bidirectional transformers for language
  understanding.
\newblock In {\em Proceedings of the 2019 conference of the North American
  chapter of the association for computational linguistics: human language
  technologies, volume 1 (long and short papers)}, pages 4171--4186, 2019.

\bibitem{brown2020language}
Tom Brown, Benjamin Mann, Nick Ryder, Melanie Subbiah, Jared~D Kaplan, Prafulla
  Dhariwal, Arvind Neelakantan, Pranav Shyam, Girish Sastry, Amanda Askell,
  et~al.
\newblock Language models are few-shot learners.
\newblock {\em Advances in neural information processing systems},
  33:1877--1901, 2020.

\bibitem{ouyang2022training}
Long Ouyang, Jeffrey Wu, Xu~Jiang, Diogo Almeida, Carroll Wainwright, Pamela
  Mishkin, Chong Zhang, Sandhini Agarwal, Katarina Slama, Alex Ray, et~al.
\newblock Training language models to follow instructions with human feedback.
\newblock {\em Advances in neural information processing systems},
  35:27730--27744, 2022.

\bibitem{raffel2020exploring}
Colin Raffel, Noam Shazeer, Adam Roberts, Katherine Lee, Sharan Narang, Michael
  Matena, Yanqi Zhou, Wei Li, and Peter~J Liu.
\newblock Exploring the limits of transfer learning with a unified text-to-text
  transformer.
\newblock {\em Journal of machine learning research}, 21(140):1--67, 2020.

\bibitem{houlsby2019parameter}
Neil Houlsby, Andrei Giurgiu, Stanislaw Jastrzebski, Bruna Morrone, Quentin
  De~Laroussilhe, Andrea Gesmundo, Mona Attariyan, and Sylvain Gelly.
\newblock Parameter-efficient transfer learning for nlp.
\newblock In {\em International conference on machine learning}, pages
  2790--2799. PMLR, 2019.

\bibitem{child2019generating}
Rewon Child, Scott Gray, Alec Radford, and Ilya Sutskever.
\newblock Generating long sequences with sparse transformers.
\newblock {\em arXiv preprint arXiv:1904.10509}, 2019.

\bibitem{mccandlish2018empirical}
Sam McCandlish, Jared Kaplan, Dario Amodei, and OpenAI~Dota Team.
\newblock An empirical model of large-batch training.
\newblock {\em arXiv preprint arXiv:1812.06162}, 2018.

\bibitem{du2022glam}
Nan Du, Yanping Huang, Andrew~M Dai, Simon Tong, Dmitry Lepikhin, Yuanzhong Xu,
  Maxim Krikun, Yanqi Zhou, Adams~Wei Yu, Orhan Firat, et~al.
\newblock Glam: Efficient scaling of language models with mixture-of-experts.
\newblock In {\em International conference on machine learning}, pages
  5547--5569. PMLR, 2022.

\bibitem{fedus2022switch}
William Fedus, Barret Zoph, and Noam Shazeer.
\newblock Switch transformers: Scaling to trillion parameter models with simple
  and efficient sparsity.
\newblock {\em Journal of Machine Learning Research}, 23(120):1--39, 2022.

\bibitem{zeng2022glm}
Aohan Zeng, Xiao Liu, Zhengxiao Du, Zihan Wang, Hanyu Lai, Ming Ding, Zhuoyi
  Yang, Yifan Xu, Wendi Zheng, Xiao Xia, et~al.
\newblock Glm-130b: An open bilingual pre-trained model.
\newblock {\em arXiv preprint arXiv:2210.02414}, 2022.

\bibitem{liu2024deepseek}
Aixin Liu, Bei Feng, Bin Wang, Bingxuan Wang, Bo~Liu, Chenggang Zhao, Chengqi
  Dengr, Chong Ruan, Damai Dai, Daya Guo, et~al.
\newblock Deepseek-v2: A strong, economical, and efficient mixture-of-experts
  language model.
\newblock {\em arXiv preprint arXiv:2405.04434}, 2024.

\bibitem{bi2024deepseek}
Xiao Bi, Deli Chen, Guanting Chen, Shanhuang Chen, Damai Dai, Chengqi Deng,
  Honghui Ding, Kai Dong, Qiushi Du, Zhe Fu, et~al.
\newblock Deepseek llm: Scaling open-source language models with longtermism.
\newblock {\em arXiv preprint arXiv:2401.02954}, 2024.

\bibitem{zhang2023cross}
Ying Zhang, Maoliang Yin, Heyong Wang, and Changchun Hua.
\newblock Cross-level multi-modal features learning with transformer for rgb-d
  object recognition.
\newblock {\em IEEE Transactions on Circuits and Systems for Video Technology},
  33(12):7121--7130, 2023.

\bibitem{joshi2020spanbert}
Mandar Joshi, Danqi Chen, Yinhan Liu, Daniel~S Weld, Luke Zettlemoyer, and Omer
  Levy.
\newblock Spanbert: Improving pre-training by representing and predicting
  spans.
\newblock {\em Transactions of the association for computational linguistics},
  8:64--77, 2020.

\bibitem{dong2019unified}
Li~Dong, Nan Yang, Wenhui Wang, Furu Wei, Xiaodong Liu, Yu~Wang, Jianfeng Gao,
  Ming Zhou, and Hsiao-Wuen Hon.
\newblock Unified language model pre-training for natural language
  understanding and generation.
\newblock {\em Advances in neural information processing systems}, 32, 2019.

\bibitem{zhou2024towards}
Pan Zhou, Xingyu Xie, Zhouchen Lin, and Shuicheng Yan.
\newblock Towards understanding convergence and generalization of adamw.
\newblock {\em IEEE transactions on pattern analysis and machine intelligence},
  2024.

\bibitem{shazeer2018adafactor}
Noam Shazeer and Mitchell Stern.
\newblock Adafactor: Adaptive learning rates with sublinear memory cost.
\newblock In {\em International Conference on Machine Learning}, pages
  4596--4604. PMLR, 2018.

\bibitem{lv2023full}
Kai Lv, Yuqing Yang, Tengxiao Liu, Qinghui Gao, Qipeng Guo, and Xipeng Qiu.
\newblock Full parameter fine-tuning for large language models with limited
  resources.
\newblock {\em arXiv preprint arXiv:2306.09782}, 2023.

\bibitem{liu2022few}
Haokun Liu, Derek Tam, Mohammed Muqeeth, Jay Mohta, Tenghao Huang, Mohit
  Bansal, and Colin~A Raffel.
\newblock Few-shot parameter-efficient fine-tuning is better and cheaper than
  in-context learning.
\newblock {\em Advances in Neural Information Processing Systems},
  35:1950--1965, 2022.

\bibitem{hu2022lora}
Edward~J Hu, Yelong Shen, Phillip Wallis, Zeyuan Allen-Zhu, Yuanzhi Li, Shean
  Wang, Lu~Wang, Weizhu Chen, et~al.
\newblock Lora: Low-rank adaptation of large language models.
\newblock {\em ICLR}, 1(2):3, 2022.

\bibitem{li2021prefix}
Xiang~Lisa Li and Percy Liang.
\newblock Prefix-tuning: Optimizing continuous prompts for generation.
\newblock {\em arXiv preprint arXiv:2101.00190}, 2021.

\bibitem{zhang2023instruction}
Shengyu Zhang, Linfeng Dong, Xiaoya Li, Sen Zhang, Xiaofei Sun, Shuhe Wang,
  Jiwei Li, Runyi Hu, Tianwei Zhang, Fei Wu, et~al.
\newblock Instruction tuning for large language models: A survey.
\newblock {\em arXiv preprint arXiv:2308.10792}, 2023.

\bibitem{kirk2023understanding}
Robert Kirk, Ishita Mediratta, Christoforos Nalmpantis, Jelena Luketina, Eric
  Hambro, Edward Grefenstette, and Roberta Raileanu.
\newblock Understanding the effects of rlhf on llm generalisation and
  diversity.
\newblock {\em arXiv preprint arXiv:2310.06452}, 2023.

\bibitem{mishra2021reframing}
Swaroop Mishra, Daniel Khashabi, Chitta Baral, Yejin Choi, and Hannaneh
  Hajishirzi.
\newblock Reframing instructional prompts to gptk's language.
\newblock {\em arXiv preprint arXiv:2109.07830}, 2021.

\bibitem{kojima2022large}
Takeshi Kojima, Shixiang~Shane Gu, Machel Reid, Yutaka Matsuo, and Yusuke
  Iwasawa.
\newblock Large language models are zero-shot reasoners.
\newblock {\em Advances in neural information processing systems},
  35:22199--22213, 2022.

\bibitem{wei2022chain}
Jason Wei, Xuezhi Wang, Dale Schuurmans, Maarten Bosma, Fei Xia, Ed~Chi, Quoc~V
  Le, Denny Zhou, et~al.
\newblock Chain-of-thought prompting elicits reasoning in large language
  models.
\newblock {\em Advances in neural information processing systems},
  35:24824--24837, 2022.

\bibitem{yang2022re3}
Kevin Yang, Yuandong Tian, Nanyun Peng, and Dan Klein.
\newblock Re3: Generating longer stories with recursive reprompting and
  revision.
\newblock {\em arXiv preprint arXiv:2210.06774}, 2022.

\bibitem{besta2024multi}
Maciej Besta, Ales Kubicek, Roman Niggli, Robert Gerstenberger, Lucas
  Weitzendorf, Mingyuan Chi, Patrick Iff, Joanna Gajda, Piotr Nyczyk,
  J{\"u}rgen M{\"u}ller, et~al.
\newblock Multi-head rag: Solving multi-aspect problems with llms.
\newblock {\em arXiv preprint arXiv:2406.05085}, 2024.

\bibitem{jeong2024adaptive}
Soyeong Jeong, Jinheon Baek, Sukmin Cho, Sung~Ju Hwang, and Jong~C Park.
\newblock Adaptive-rag: Learning to adapt retrieval-augmented large language
  models through question complexity.
\newblock {\em arXiv preprint arXiv:2403.14403}, 2024.

\bibitem{sawarkar2024blended}
Kunal Sawarkar, Abhilasha Mangal, and Shivam~Raj Solanki.
\newblock Blended rag: Improving rag (retriever-augmented generation) accuracy
  with semantic search and hybrid query-based retrievers.
\newblock In {\em 2024 IEEE 7th International Conference on Multimedia
  Information Processing and Retrieval (MIPR)}, pages 155--161. IEEE, 2024.

\bibitem{asai2023self}
Akari Asai, Zeqiu Wu, Yizhong Wang, Avirup Sil, and Hannaneh Hajishirzi.
\newblock Self-rag: Learning to retrieve, generate, and critique through
  self-reflection.
\newblock In {\em The Twelfth International Conference on Learning
  Representations}, 2023.

\bibitem{yang2024rag}
Diji Yang, Jinmeng Rao, Kezhen Chen, Xiaoyuan Guo, Yawen Zhang, Jie Yang, and
  Yi~Zhang.
\newblock Im-rag: Multi-round retrieval-augmented generation through learning
  inner monologues.
\newblock In {\em Proceedings of the 47th International ACM SIGIR Conference on
  Research and Development in Information Retrieval}, pages 730--740, 2024.

\bibitem{ding2023task}
Yan Ding, Xiaohan Zhang, Chris Paxton, and Shiqi Zhang.
\newblock Task and motion planning with large language models for object
  rearrangement.
\newblock In {\em 2023 IEEE/RSJ International Conference on Intelligent Robots
  and Systems (IROS)}, pages 2086--2092. IEEE, 2023.

\bibitem{singh2023progprompt}
Ishika Singh, Valts Blukis, Arsalan Mousavian, Ankit Goyal, Danfei Xu, Jonathan
  Tremblay, Dieter Fox, Jesse Thomason, and Animesh Garg.
\newblock Progprompt: Generating situated robot task plans using large language
  models.
\newblock In {\em 2023 IEEE International Conference on Robotics and Automation
  (ICRA)}, pages 11523--11530. IEEE, 2023.

\bibitem{silver2024generalized}
Tom Silver, Soham Dan, Kavitha Srinivas, Joshua~B Tenenbaum, Leslie Kaelbling,
  and Michael Katz.
\newblock Generalized planning in pddl domains with pretrained large language
  models.
\newblock In {\em Proceedings of the AAAI conference on artificial
  intelligence}, volume~38, pages 20256--20264, 2024.

\bibitem{li2024fine}
Xiaodong Li, Guohui Tian, and Yongcheng Cui.
\newblock Fine-grained task planning for service robots based on object
  ontology knowledge via large language models.
\newblock {\em IEEE Robotics and Automation Letters}, 2024.

\bibitem{wu2023embodied}
Zhenyu Wu, Ziwei Wang, Xiuwei Xu, Jiwen Lu, and Haibin Yan.
\newblock Embodied task planning with large language models.
\newblock {\em arXiv preprint arXiv:2307.01848}, 2023.

\bibitem{ren2023robots}
Allen~Z Ren, Anushri Dixit, Alexandra Bodrova, Sumeet Singh, Stephen Tu, Noah
  Brown, Peng Xu, Leila Takayama, Fei Xia, Jake Varley, et~al.
\newblock Robots that ask for help: Uncertainty alignment for large language
  model planners.
\newblock {\em arXiv preprint arXiv:2307.01928}, 2023.

\bibitem{liu2024delta}
Yuchen Liu, Luigi Palmieri, Sebastian Koch, Ilche Georgievski, and Marco
  Aiello.
\newblock Delta: Decomposed efficient long-term robot task planning using large
  language models.
\newblock {\em arXiv preprint arXiv:2404.03275}, 2024.

\bibitem{kannan2024smart}
Shyam~Sundar Kannan, Vishnunandan~LN Venkatesh, and Byung-Cheol Min.
\newblock Smart-llm: Smart multi-agent robot task planning using large language
  models.
\newblock In {\em 2024 IEEE/RSJ International Conference on Intelligent Robots
  and Systems (IROS)}, pages 12140--12147. IEEE, 2024.

\bibitem{liu2023llm+}
Bo~Liu, Yuqian Jiang, Xiaohan Zhang, Qiang Liu, Shiqi Zhang, Joydeep Biswas,
  and Peter Stone.
\newblock Llm+ p: Empowering large language models with optimal planning
  proficiency.
\newblock {\em arXiv preprint arXiv:2304.11477}, 2023.

\bibitem{kawabe2024task}
Tomoya Kawabe, Tatsushi Nishi, Ziang Liu, and Tomofumi Fujiwara.
\newblock Task planning for robot manipulator using natural language task input
  with large language models.
\newblock In {\em 2024 IEEE 20th International Conference on Automation Science
  and Engineering (CASE)}, pages 3484--3489. IEEE, 2024.

\bibitem{chen2025extendable}
Chang Chen, Hany Hamed, Doojin Baek, Taegu Kang, Yoshua Bengio, and Sungjin
  Ahn.
\newblock Extendable long-horizon planning via hierarchical multiscale
  diffusion.
\newblock {\em arXiv preprint arXiv:2503.20102}, 2025.

\bibitem{erdogan2025plan}
Lutfi~Eren Erdogan, Nicholas Lee, Sehoon Kim, Suhong Moon, Hiroki Furuta,
  Gopala Anumanchipalli, Kurt Keutzer, and Amir Gholami.
\newblock Plan-and-act: Improving planning of agents for long-horizon tasks.
\newblock {\em arXiv preprint arXiv:2503.09572}, 2025.

\bibitem{cao2024llm}
Hong Cao, Rong Ma, Yanlong Zhai, and Jun Shen.
\newblock Llm-collab: a framework for enhancing task planning via
  chain-of-thought and multi-agent collaboration.
\newblock {\em Applied Computing and Intelligence}, 4(2):328--348, 2024.

\bibitem{zhen2023robot}
Yue Zhen, Sheng Bi, Lu~Xing-tong, Pan Wei-qin, Shi Hai-peng, Chen Zi-rui, and
  Fang Yi-shu.
\newblock Robot task planning based on large language model representing
  knowledge with directed graph structures.
\newblock {\em arXiv preprint arXiv:2306.05171}, 2023.

\bibitem{park2023clara}
Jeongeun Park, Seungwon Lim, Joonhyung Lee, Sangbeom Park, Minsuk Chang,
  Youngjae Yu, and Sungjoon Choi.
\newblock Clara: classifying and disambiguating user commands for reliable
  interactive robotic agents.
\newblock {\em IEEE Robotics and Automation Letters}, 9(2):1059--1066, 2023.

\bibitem{ong2024simple}
Hyobin Ong, Youngwoo Yoon, Jaewoo Choi, and Minsu Jang.
\newblock A simple baseline for uncertainty-aware language-oriented task
  planner for embodied agents.
\newblock In {\em 2024 21st International Conference on Ubiquitous Robots
  (UR)}, pages 677--682. IEEE, 2024.

\bibitem{wang2024llm}
Ruoyu Wang, Zhipeng Yang, Zinan Zhao, Xinyan Tong, Zhi Hong, and Kun Qian.
\newblock Llm-based robot task planning with exceptional handling for general
  purpose service robots.
\newblock In {\em 2024 43rd Chinese Control Conference (CCC)}, pages
  4439--4444. IEEE, 2024.

\bibitem{mu2023kggpt}
Zonghao Mu, Wenyu Zhao, Yue Yin, Xiangming Xi, Wei Song, Jianjun Gu, and
  Shiqiang Zhu.
\newblock Kggpt: empowering robots with openai’s chatgpt and knowledge graph.
\newblock In {\em International Conference on Intelligent Robotics and
  Applications}, pages 340--351. Springer, 2023.

\bibitem{ding2023integrating}
Yan Ding, Xiaohan Zhang, Saeid Amiri, Nieqing Cao, Hao Yang, Andy Kaminski,
  Chad Esselink, and Shiqi Zhang.
\newblock Integrating action knowledge and llms for task planning and situation
  handling in open worlds.
\newblock {\em Autonomous Robots}, 47(8):981--997, 2023.

\bibitem{zhang2025zisvfm}
Ying Zhang, Maoliang Yin, Wenfu Bi, Haibao Yan, Shaohan Bian, Cui-Hua Zhang,
  and Changchun Hua.
\newblock Zisvfm: Zero-shot object instance segmentation in indoor robotic
  environments with vision foundation models.
\newblock {\em IEEE Transactions on Robotics}, 41:1568--1580, 2025.

\bibitem{tang20253d}
Guoqin Tang, Qingxuan Jia, Zeyuan Huang, Gang Chen, Ning Ji, and Zhipeng Yao.
\newblock 3d-grounded vision-language framework for robotic task planning:
  Automated prompt synthesis and supervised reasoning.
\newblock {\em arXiv preprint arXiv:2502.08903}, 2025.

\bibitem{ni2024grid}
Zhe Ni, Xiaoxin Deng, Cong Tai, Xinyue Zhu, Qinghongbing Xie, Weihang Huang,
  Xiang Wu, and Long Zeng.
\newblock Grid: Scene-graph-based instruction-driven robotic task planning.
\newblock In {\em 2024 IEEE/RSJ International Conference on Intelligent Robots
  and Systems (IROS)}, pages 13765--13772. IEEE, 2024.

\bibitem{ekpo2024verigraph}
Daniel Ekpo, Mara Levy, Saksham Suri, Chuong Huynh, and Abhinav Shrivastava.
\newblock Verigraph: Scene graphs for execution verifiable robot planning.
\newblock {\em arXiv preprint arXiv:2411.10446}, 2024.

\bibitem{mei2024replanvlm}
Aoran Mei, Guo-Niu Zhu, Huaxiang Zhang, and Zhongxue Gan.
\newblock Replanvlm: Replanning robotic tasks with visual language models.
\newblock {\em IEEE Robotics and Automation Letters}, 2024.

\bibitem{huang2023visual}
Chenguang Huang, Oier Mees, Andy Zeng, and Wolfram Burgard.
\newblock Visual language maps for robot navigation.
\newblock In {\em 2023 IEEE International Conference on Robotics and Automation
  (ICRA)}, pages 10608--10615. IEEE, 2023.

\bibitem{shirai2024vision}
Keisuke Shirai, Cristian~C Beltran-Hernandez, Masashi Hamaya, Atsushi
  Hashimoto, Shohei Tanaka, Kento Kawaharazuka, Kazutoshi Tanaka, Yoshitaka
  Ushiku, and Shinsuke Mori.
\newblock Vision-language interpreter for robot task planning.
\newblock In {\em 2024 IEEE International Conference on Robotics and Automation
  (ICRA)}, pages 2051--2058. IEEE, 2024.

\bibitem{luan2024enhancing}
Zhirong Luan, Yujun Lai, Rundong Huang, Shuanghao Bai, Yuedi Zhang, Haoran
  Zhang, and Qian Wang.
\newblock Enhancing robot task planning and execution through multi-layer large
  language models.
\newblock {\em Sensors}, 24(5):1687, 2024.

\bibitem{huang2024combining}
Jiayang Huang, Christian Limberg, Syed Muhammad~Nashit Arshad, Qifeng Zhang,
  and Qiang Li.
\newblock Combining vlm and llm for enhanced semantic object perception in
  robotic handover tasks.
\newblock In {\em 2024 WRC Symposium on Advanced Robotics and Automation (WRC
  SARA)}, pages 135--140. IEEE, 2024.

\bibitem{zhang2025gptarm}
Jiaqi Zhang, Zinan Wang, Jiaxin Lai, and Hongfei Wang.
\newblock Gptarm: An autonomous task planning manipulator grasping system based
  on vision--language models.
\newblock {\em Machines}, 13(3):247, 2025.

\bibitem{wake2024gpt}
Naoki Wake, Atsushi Kanehira, Kazuhiro Sasabuchi, Jun Takamatsu, and Katsushi
  Ikeuchi.
\newblock Gpt-4v (ision) for robotics: Multimodal task planning from human
  demonstration.
\newblock {\em IEEE Robotics and Automation Letters}, 2024.

\bibitem{liu2024vision}
Sichao Liu, Jianjing Zhang, Robert~X Gao, Xi~Vincent Wang, and Lihui Wang.
\newblock Vision-language model-driven scene understanding and robotic object
  manipulation.
\newblock In {\em 2024 IEEE 20th International Conference on Automation Science
  and Engineering (CASE)}, pages 21--26. IEEE, 2024.

\bibitem{liu2022behavior}
Shaopeng Liu, Guohui Tian, Xuyang Shao, and Shuo Liu.
\newblock Behavior cloning-based robot active object detection with
  automatically generated data and revision method.
\newblock {\em IEEE Transactions on Robotics}, 39(1):665--680, 2022.

\bibitem{mo2021mutual}
Shaocong Mo, Ming Cai, Lanfen Lin, Ruofeng Tong, Qingqing Chen, Fang Wang,
  Hongjie Hu, Yutaro Iwamoto, Xian-Hua Han, and Yen-Wei Chen.
\newblock Mutual information-based graph co-attention networks for multimodal
  prior-guided magnetic resonance imaging segmentation.
\newblock {\em IEEE Transactions on Circuits and Systems for Video Technology},
  32(5):2512--2526, 2021.

\bibitem{liu2021toward}
An-An Liu, Hongshuo Tian, Ning Xu, Weizhi Nie, Yongdong Zhang, and Mohan
  Kankanhalli.
\newblock Toward region-aware attention learning for scene graph generation.
\newblock {\em IEEE Transactions on Neural Networks and Learning Systems},
  33(12):7655--7666, 2021.

\bibitem{zhang2022sequential}
Mengyang Zhang, Guohui Tian, Ying Zhang, and Hong Liu.
\newblock Sequential learning for ingredient recognition from images.
\newblock {\em IEEE Transactions on Circuits and Systems for Video Technology},
  33(5):2162--2175, 2022.

\bibitem{xie2022knowledge}
Jiayuan Xie, Wenhao Fang, Yi~Cai, Qingbao Huang, and Qing Li.
\newblock Knowledge-based visual question generation.
\newblock {\em IEEE Transactions on Circuits and Systems for Video Technology},
  32(11):7547--7558, 2022.

\bibitem{hao2024embosr}
Yu~Hao, Fan Yang, Nicholas Fang, and Yu-Shen Liu.
\newblock Embosr: Embodied spatial reasoning for enhanced situated question
  answering in 3d scenes.
\newblock In {\em 2024 IEEE/RSJ International Conference on Intelligent Robots
  and Systems (IROS)}, pages 9811--9816. IEEE, 2024.

\bibitem{cheng2024spatialrgpt}
An-Chieh Cheng, Hongxu Yin, Yang Fu, Qiushan Guo, Ruihan Yang, Jan Kautz,
  Xiaolong Wang, and Sifei Liu.
\newblock Spatialrgpt: Grounded spatial reasoning in vision language models.
\newblock {\em arXiv preprint arXiv:2406.01584}, 2024.

\bibitem{yang2021solver}
Jingyuan Yang, Xinbo Gao, Leida Li, Xiumei Wang, and Jinshan Ding.
\newblock Solver: Scene-object interrelated visual emotion reasoning network.
\newblock {\em IEEE Transactions on Image Processing}, 30:8686--8701, 2021.

\bibitem{gu2024interactive}
Weiwei Gu, Anant Sah, and Nakul Gopalan.
\newblock Interactive visual task learning for robots.
\newblock In {\em Proceedings of the AAAI Conference on Artificial
  Intelligence}, volume~38, pages 10297--10305, 2024.

\bibitem{mascaro2024scene}
Ruben Mascaro and Margarita Chli.
\newblock Scene representations for robotic spatial perception.
\newblock {\em Annual Review of Control, Robotics, and Autonomous Systems}, 8,
  2024.

\bibitem{senior2025graph}
Henry Senior, Gregory Slabaugh, Shanxin Yuan, and Luca Rossi.
\newblock Graph neural networks in vision-language image understanding: a
  survey.
\newblock {\em The Visual Computer}, 41(1):491--516, 2025.

\bibitem{jiao2022sequential}
Ziyuan Jiao, Yida Niu, Zeyu Zhang, Song-Chun Zhu, Yixin Zhu, and Hangxin Liu.
\newblock Sequential manipulation planning on scene graph.
\newblock In {\em 2022 IEEE/RSJ International Conference on Intelligent Robots
  and Systems (IROS)}, pages 8203--8210. IEEE, 2022.

\bibitem{kenfack2020robotvqa}
Franklin~Kenghagho Kenfack, Feroz~Ahmed Siddiky, Ferenc Balint-Benczedi, and
  Michael Beetz.
\newblock Robotvqa—a scene-graph-and deep-learning-based visual question
  answering system for robot manipulation.
\newblock In {\em 2020 IEEE/RSJ International Conference on Intelligent Robots
  and Systems (IROS)}, pages 9667--9674. IEEE, 2020.

\bibitem{wang2024cog}
Shaowei Wang, Lingling Zhang, Longji Zhu, Tao Qin, Kim-Hui Yap, Xinyu Zhang,
  and Jun Liu.
\newblock Cog-dqa: Chain-of-guiding learning with large language models for
  diagram question answering.
\newblock In {\em Proceedings of the IEEE/CVF Conference on Computer Vision and
  Pattern Recognition}, pages 13969--13979, 2024.

\bibitem{elizalde2023clap}
Benjamin Elizalde, Soham Deshmukh, Mahmoud Al~Ismail, and Huaming Wang.
\newblock Clap learning audio concepts from natural language supervision.
\newblock In {\em ICASSP 2023-2023 IEEE International Conference on Acoustics,
  Speech and Signal Processing (ICASSP)}, pages 1--5. IEEE, 2023.

\bibitem{guzhov2022audioclip}
Andrey Guzhov, Federico Raue, J{\"o}rn Hees, and Andreas Dengel.
\newblock Audioclip: Extending clip to image, text and audio.
\newblock In {\em ICASSP 2022-2022 IEEE International Conference on Acoustics,
  Speech and Signal Processing (ICASSP)}, pages 976--980. IEEE, 2022.

\bibitem{kong2024audio}
Zhifeng Kong, Arushi Goel, Rohan Badlani, Wei Ping, Rafael Valle, and Bryan
  Catanzaro.
\newblock Audio flamingo: A novel audio language model with few-shot learning
  and dialogue abilities.
\newblock {\em arXiv preprint arXiv:2402.01831}, 2024.

\bibitem{chen2024salm}
Zhehuai Chen, He~Huang, Andrei Andrusenko, Oleksii Hrinchuk, Krishna~C Puvvada,
  Jason Li, Subhankar Ghosh, Jagadeesh Balam, and Boris Ginsburg.
\newblock Salm: Speech-augmented language model with in-context learning for
  speech recognition and translation.
\newblock In {\em ICASSP 2024-2024 IEEE International Conference on Acoustics,
  Speech and Signal Processing (ICASSP)}, pages 13521--13525. IEEE, 2024.

\bibitem{rubenstein2023audiopalm}
Paul~K Rubenstein, Chulayuth Asawaroengchai, Duc~Dung Nguyen, Ankur Bapna,
  Zal{\'a}n Borsos, F{\'e}lix de~Chaumont Quitry, Peter Chen, Dalia~El Badawy,
  Wei Han, Eugene Kharitonov, et~al.
\newblock Audiopalm: A large language model that can speak and listen.
\newblock {\em arXiv preprint arXiv:2306.12925}, 2023.

\bibitem{wang2024can}
Siyin Wang, Chao-Han Yang, Ji~Wu, and Chao Zhang.
\newblock Can whisper perform speech-based in-context learning?
\newblock In {\em ICASSP 2024-2024 IEEE International Conference on Acoustics,
  Speech and Signal Processing (ICASSP)}, pages 13421--13425. IEEE, 2024.

\bibitem{deshmukh2023pengi}
Soham Deshmukh, Benjamin Elizalde, Rita Singh, and Huaming Wang.
\newblock Pengi: An audio language model for audio tasks.
\newblock {\em Advances in Neural Information Processing Systems},
  36:18090--18108, 2023.

\bibitem{wu2022wav2clip}
Ho-Hsiang Wu, Prem Seetharaman, Kundan Kumar, and Juan~Pablo Bello.
\newblock Wav2clip: Learning robust audio representations from clip.
\newblock In {\em ICASSP 2022-2022 IEEE International Conference on Acoustics,
  Speech and Signal Processing (ICASSP)}, pages 4563--4567. IEEE, 2022.

\bibitem{gong2023listen}
Yuan Gong, Hongyin Luo, Alexander~H Liu, Leonid Karlinsky, and James Glass.
\newblock Listen, think, and understand.
\newblock {\em arXiv preprint arXiv:2305.10790}, 2023.

\bibitem{ghosh2024gama}
Sreyan Ghosh, Sonal Kumar, Ashish Seth, Chandra Kiran~Reddy Evuru, Utkarsh
  Tyagi, S~Sakshi, Oriol Nieto, Ramani Duraiswami, and Dinesh Manocha.
\newblock Gama: A large audio-language model with advanced audio understanding
  and complex reasoning abilities.
\newblock {\em arXiv preprint arXiv:2406.11768}, 2024.

\bibitem{sun2024beyond}
Xingpeng Sun, Haoming Meng, Souradip Chakraborty, Amrit~Singh Bedi, and Aniket
  Bera.
\newblock Beyond text: Utilizing vocal cues to improve decision making in llms
  for robot navigation tasks.
\newblock {\em arXiv preprint arXiv:2402.03494}, 2024.

\bibitem{anil2023palm}
Rohan Anil, Andrew~M Dai, Orhan Firat, Melvin Johnson, Dmitry Lepikhin,
  Alexandre Passos, Siamak Shakeri, Emanuel Taropa, Paige Bailey, Zhifeng Chen,
  et~al.
\newblock Palm 2 technical report.
\newblock {\em arXiv preprint arXiv:2305.10403}, 2023.

\bibitem{borsos2023audiolm}
Zal{\'a}n Borsos, Rapha{\"e}l Marinier, Damien Vincent, Eugene Kharitonov,
  Olivier Pietquin, Matt Sharifi, Dominik Roblek, Olivier Teboul, David
  Grangier, Marco Tagliasacchi, et~al.
\newblock Audiolm: a language modeling approach to audio generation.
\newblock {\em IEEE/ACM transactions on audio, speech, and language
  processing}, 31:2523--2533, 2023.

\bibitem{sakshi2024mmau}
S~Sakshi, Utkarsh Tyagi, Sonal Kumar, Ashish Seth, Ramaneswaran Selvakumar,
  Oriol Nieto, Ramani Duraiswami, Sreyan Ghosh, and Dinesh Manocha.
\newblock Mmau: A massive multi-task audio understanding and reasoning
  benchmark.
\newblock {\em arXiv preprint arXiv:2410.19168}, 2024.

\bibitem{salewski2023zero}
Leonard Salewski, Stefan Fauth, A~Koepke, and Zeynep Akata.
\newblock Zero-shot audio captioning with audio-language model guidance and
  audio context keywords.
\newblock {\em arXiv preprint arXiv:2311.08396}, 2023.

\bibitem{hu2024listen}
Yuchen Hu, Chen Chen, Chengwei Qin, Qiushi Zhu, Eng~Siong Chng, and Ruizhe Li.
\newblock Listen again and choose the right answer: A new paradigm for
  automatic speech recognition with large language models.
\newblock {\em arXiv preprint arXiv:2405.10025}, 2024.

\bibitem{ji2025learning}
Chao Ji, Diyuan Liu, Wei Gao, and Shiwu Zhang.
\newblock Learning-based locomotion control fusing multimodal perception for a
  bipedal humanoid robot.
\newblock {\em Biomimetic Intelligence and Robotics}, page 100213, 2025.

\bibitem{alayrac2022flamingo}
Jean-Baptiste Alayrac, Jeff Donahue, Pauline Luc, Antoine Miech, Iain Barr,
  Yana Hasson, Karel Lenc, Arthur Mensch, Katherine Millican, Malcolm Reynolds,
  et~al.
\newblock Flamingo: a visual language model for few-shot learning.
\newblock {\em Advances in neural information processing systems},
  35:23716--23736, 2022.

\bibitem{li2023blip}
Junnan Li, Dongxu Li, Silvio Savarese, and Steven Hoi.
\newblock Blip-2: Bootstrapping language-image pre-training with frozen image
  encoders and large language models.
\newblock In {\em International conference on machine learning}, pages
  19730--19742. PMLR, 2023.

\bibitem{yang2023mm}
Zhengyuan Yang, Linjie Li, Jianfeng Wang, Kevin Lin, Ehsan Azarnasab, Faisal
  Ahmed, Zicheng Liu, Ce~Liu, Michael Zeng, and Lijuan Wang.
\newblock Mm-react: Prompting chatgpt for multimodal reasoning and action.
\newblock {\em arXiv preprint arXiv:2303.11381}, 2023.

\bibitem{zhang2023llama}
Renrui Zhang, Jiaming Han, Chris Liu, Peng Gao, Aojun Zhou, Xiangfei Hu, Shilin
  Yan, Pan Lu, Hongsheng Li, and Yu~Qiao.
\newblock Llama-adapter: Efficient fine-tuning of language models with
  zero-init attention.
\newblock {\em arXiv preprint arXiv:2303.16199}, 2023.

\bibitem{zhu2023minigpt}
Deyao Zhu, Jun Chen, Xiaoqian Shen, Xiang Li, and Mohamed Elhoseiny.
\newblock Minigpt-4: Enhancing vision-language understanding with advanced
  large language models.
\newblock {\em arXiv preprint arXiv:2304.10592}, 2023.

\bibitem{liu2023visual}
Haotian Liu, Chunyuan Li, Qingyang Wu, and Yong~Jae Lee.
\newblock Visual instruction tuning.
\newblock {\em Advances in neural information processing systems},
  36:34892--34916, 2023.

\bibitem{liu2024improved}
Haotian Liu, Chunyuan Li, Yuheng Li, and Yong~Jae Lee.
\newblock Improved baselines with visual instruction tuning.
\newblock In {\em Proceedings of the IEEE/CVF Conference on Computer Vision and
  Pattern Recognition}, pages 26296--26306, 2024.

\bibitem{peng2023kosmos}
Zhiliang Peng, Wenhui Wang, Li~Dong, Yaru Hao, Shaohan Huang, Shuming Ma, and
  Furu Wei.
\newblock Kosmos-2: Grounding multimodal large language models to the world.
\newblock {\em arXiv preprint arXiv:2306.14824}, 2023.

\bibitem{wang2024qwen2}
Peng Wang, Shuai Bai, Sinan Tan, Shijie Wang, Zhihao Fan, Jinze Bai, Keqin
  Chen, Xuejing Liu, Jialin Wang, Wenbin Ge, et~al.
\newblock Qwen2-vl: Enhancing vision-language model's perception of the world
  at any resolution.
\newblock {\em arXiv preprint arXiv:2409.12191}, 2024.

\bibitem{ye2023mplug}
Qinghao Ye, Haiyang Xu, Guohai Xu, Jiabo Ye, Ming Yan, Yiyang Zhou, Junyang
  Wang, Anwen Hu, Pengcheng Shi, Yaya Shi, et~al.
\newblock mplug-owl: Modularization empowers large language models with
  multimodality.
\newblock {\em arXiv preprint arXiv:2304.14178}, 2023.

\bibitem{li2023mimic}
Bo~Li, Yuanhan Zhang, Liangyu Chen, Jinghao Wang, Fanyi Pu, Jingkang Yang,
  Chunyuan Li, and Ziwei Liu.
\newblock Mimic-it: Multi-modal in-context instruction tuning.
\newblock {\em arXiv preprint arXiv:2306.05425}, 2023.

\bibitem{qi2024cogcom}
Ji~Qi, Ming Ding, Weihan Wang, Yushi Bai, Qingsong Lv, Wenyi Hong, Bin Xu, Lei
  Hou, Juanzi Li, Yuxiao Dong, et~al.
\newblock Cogcom: Train large vision-language models diving into details
  through chain of manipulations.
\newblock {\em arXiv preprint arXiv:2402.04236}, 2024.

\bibitem{yan2024vigor}
Siming Yan, Min Bai, Weifeng Chen, Xiong Zhou, Qixing Huang, and Li~Erran Li.
\newblock Vigor: Improving visual grounding of large vision language models
  with fine-grained reward modeling.
\newblock In {\em European Conference on Computer Vision}, pages 37--53.
  Springer, 2024.

\bibitem{radford2021learning}
Alec Radford, Jong~Wook Kim, Chris Hallacy, Aditya Ramesh, Gabriel Goh,
  Sandhini Agarwal, Girish Sastry, Amanda Askell, Pamela Mishkin, Jack Clark,
  et~al.
\newblock Learning transferable visual models from natural language
  supervision.
\newblock In {\em International conference on machine learning}, pages
  8748--8763. PmLR, 2021.

\bibitem{li2024omnibench}
Yizhi Li, Ge~Zhang, Yinghao Ma, Ruibin Yuan, Kang Zhu, Hangyu Guo, Yiming
  Liang, Jiaheng Liu, Zekun Wang, Jian Yang, et~al.
\newblock Omnibench: Towards the future of universal omni-language models.
\newblock {\em arXiv preprint arXiv:2409.15272}, 2024.

\bibitem{huang2023modality}
Kai Huang, Boyuan Yang, and Wei Gao.
\newblock Modality plug-and-play: Elastic modality adaptation in multimodal
  llms for embodied ai.
\newblock {\em arXiv preprint arXiv:2312.07886}, 2023.

\bibitem{zhu2024llmbind}
Bin Zhu, Munan Ning, Peng Jin, Bin Lin, Jinfa Huang, Qi~Song, Junwu Zhang,
  Zhenyu Tang, Mingjun Pan, Xing Zhou, et~al.
\newblock Llmbind: A unified modality-task integration framework.
\newblock {\em arXiv preprint arXiv:2402.14891}, 2024.

\bibitem{zhang2024respllm}
Yuwei Zhang, Tong Xia, Aaqib Saeed, and Cecilia Mascolo.
\newblock Respllm: Unifying audio and text with multimodal llms for generalized
  respiratory health prediction.
\newblock {\em arXiv preprint arXiv:2410.05361}, 2024.

\bibitem{ni2024design}
Mingze Ni, Gang Xu, Hongsen Li, Zhaoming Luo, Limin Pang, and Binchao Yu.
\newblock Design of a service robot system based on a multimodal large model.
\newblock In {\em 2024 6th International Symposium on Robotics \& Intelligent
  Manufacturing Technology (ISRIMT)}, pages 81--86. IEEE, 2024.

\bibitem{chung2024empowering}
Tong~Lee Chung, Jianxin Pang, and Jun Cheng.
\newblock Empowering robots with multimodal language models for task planning
  with interaction.
\newblock In {\em 2024 IEEE 14th International Symposium on Chinese Spoken
  Language Processing (ISCSLP)}, pages 358--362. IEEE, 2024.

\bibitem{jiang2023vima}
Yunfan Jiang, Agrim Gupta, Zichen Zhang, Guanzhi Wang, Yongqiang Dou, Yanjun
  Chen, Li~Fei-Fei, Anima Anandkumar, Yuke Zhu, and Linxi Fan.
\newblock Vima: Robot manipulation with multimodal prompts.
\newblock 2023.

\bibitem{huang2022inner}
Wenlong Huang, Fei Xia, Ted Xiao, Harris Chan, Jacky Liang, Pete Florence, Andy
  Zeng, Jonathan Tompson, Igor Mordatch, Yevgen Chebotar, et~al.
\newblock Inner monologue: Embodied reasoning through planning with language
  models.
\newblock {\em arXiv preprint arXiv:2207.05608}, 2022.

\bibitem{liu2024enhancing}
Yang Liu, Yanchao Zhao, Weichao Guo, Xinjun Sheng, and Han Ding.
\newblock Enhancing household service robots with a dual-arm mobile manipulator
  and multimodal large language models.
\newblock In {\em 2024 IEEE International Conference on Robotics and
  Biomimetics (ROBIO)}, pages 1815--1820. IEEE, 2024.

\bibitem{wang2024large}
Jiaqi Wang, Enze Shi, Huawen Hu, Chong Ma, Yiheng Liu, Xuhui Wang, Yincheng
  Yao, Xuan Liu, Bao Ge, and Shu Zhang.
\newblock Large language models for robotics: Opportunities, challenges, and
  perspectives.
\newblock {\em Journal of Automation and Intelligence}, 2024.

\bibitem{driess2023palm}
Danny Driess, Fei Xia, Mehdi~SM Sajjadi, Corey Lynch, Aakanksha Chowdhery,
  Ayzaan Wahid, Jonathan Tompson, Quan Vuong, Tianhe Yu, Wenlong Huang, et~al.
\newblock Palm-e: An embodied multimodal language model.
\newblock 2023.

\bibitem{kambara2024human}
Motonari Kambara, Chiori Hori, Komei Sugiura, Kei Ota, Devesh~K Jha, Sameer
  Khurana, Siddarth Jain, Radu Corcodel, Diego Romeres, and Jonathan Le~Roux.
\newblock Human action understanding-based robot planning using multimodal llm.
\newblock In {\em IEEE International Conference on Robotics and Automation
  (ICRA) Workshop}, 2024.

\bibitem{lai2025nmm}
Yuzhi Lai, Shenghai Yuan, Youssef Nassar, Mingyu Fan, Atmaraaj Gopal, Arihiro
  Yorita, Naoyuki Kubota, and Matthias R{\"a}tsch.
\newblock Nmm-hri: Natural multi-modal human-robot interaction with voice and
  deictic posture via large language model.
\newblock {\em arXiv preprint arXiv:2501.00785}, 2025.

\bibitem{tian11task}
Guohui Tian$^1$, Jian Jiang, and Shanmei Wang.
\newblock Task reasoning of service robots with fused.
\newblock In {\em Proceedings of the 3rd International Conference on Machine
  Learning, Cloud Computing and Intelligent Mining (MLCCIM2024): Volume 1},
  page 347. Springer Nature.

\bibitem{team2025gemini}
Gemini~Robotics Team, Saminda Abeyruwan, Joshua Ainslie, Jean-Baptiste Alayrac,
  Montserrat~Gonzalez Arenas, Travis Armstrong, Ashwin Balakrishna, Robert
  Baruch, Maria Bauza, Michiel Blokzijl, et~al.
\newblock Gemini robotics: Bringing ai into the physical world.
\newblock {\em arXiv preprint arXiv:2503.20020}, 2025.

\bibitem{bjorck2025gr00t}
Johan Bjorck, Fernando Casta{\~n}eda, Nikita Cherniadev, Xingye Da, Runyu Ding,
  Linxi Fan, Yu~Fang, Dieter Fox, Fengyuan Hu, Spencer Huang, et~al.
\newblock Gr00t n1: An open foundation model for generalist humanoid robots.
\newblock {\em arXiv preprint arXiv:2503.14734}, 2025.

\bibitem{team2024gemini}
Gemini Team, Petko Georgiev, Ving~Ian Lei, Ryan Burnell, Libin Bai, Anmol
  Gulati, Garrett Tanzer, Damien Vincent, Zhufeng Pan, Shibo Wang, et~al.
\newblock Gemini 1.5: Unlocking multimodal understanding across millions of
  tokens of context.
\newblock {\em arXiv preprint arXiv:2403.05530}, 2024.

\bibitem{achiam2023gpt}
Josh Achiam, Steven Adler, Sandhini Agarwal, Lama Ahmad, Ilge Akkaya,
  Florencia~Leoni Aleman, Diogo Almeida, Janko Altenschmidt, Sam Altman,
  Shyamal Anadkat, et~al.
\newblock Gpt-4 technical report.
\newblock {\em arXiv preprint arXiv:2303.08774}, 2023.

\bibitem{grattafiori2024llama}
Aaron Grattafiori, Abhimanyu Dubey, Abhinav Jauhri, Abhinav Pandey, Abhishek
  Kadian, Ahmad Al-Dahle, Aiesha Letman, Akhil Mathur, Alan Schelten, Alex
  Vaughan, et~al.
\newblock The llama 3 herd of models.
\newblock {\em arXiv preprint arXiv:2407.21783}, 2024.

\bibitem{zheng2025llm}
Yangqing Zheng, Shunqi Mao, Dingxin Zhang, and Weidong Cai.
\newblock Llm-enhanced rapid-reflex async-reflect embodied agent for real-time
  decision-making in dynamically changing environments.
\newblock {\em arXiv preprint arXiv:2506.07223}, 2025.

\bibitem{sun2025llm}
Chuanneng Sun, Songjun Huang, and Dario Pompili.
\newblock Llm-based multi-agent decision-making: Challenges and future
  directions.
\newblock {\em IEEE Robotics and Automation Letters}, 2025.

\bibitem{chen2025robogpt}
Yaran Chen, Wenbo Cui, Yuanwen Chen, Mining Tan, Xinyao Zhang, Jinrui Liu,
  Haoran Li, Dongbin Zhao, and He~Wang.
\newblock Robogpt: an llm-based long-term decision-making embodied agent for
  instruction following tasks.
\newblock {\em IEEE Transactions on Cognitive and Developmental Systems}, 2025.

\bibitem{saha2025vision}
Shaibal Saha and Lanyu Xu.
\newblock Vision transformers on the edge: A comprehensive survey of model
  compression and acceleration strategies.
\newblock {\em Neurocomputing}, page 130417, 2025.

\bibitem{sarridis2025indistill}
Ioannis Sarridis, Christos Koutlis, Giorgos Kordopatis-Zilos, Ioannis
  Kompatsiaris, and Symeon Papadopoulos.
\newblock Indistill: Information flow-preserving knowledge distillation for
  model compression.
\newblock In {\em 2025 IEEE/CVF Winter Conference on Applications of Computer
  Vision (WACV)}, pages 9033--9042. IEEE, 2025.

\bibitem{liu2025enhancing}
Chang Liu and Jun Zhao.
\newblock Enhancing stability and resource efficiency in llm training for
  edge-assisted mobile systems.
\newblock {\em IEEE Transactions on Mobile Computing}, 2025.

\bibitem{friel2023chainpoll}
Robert Friel and Atindriyo Sanyal.
\newblock Chainpoll: A high efficacy method for llm hallucination detection.
\newblock {\em arXiv preprint arXiv:2310.18344}, 2023.

\bibitem{martino2023knowledge}
Ariana Martino, Michael Iannelli, and Coleen Truong.
\newblock Knowledge injection to counter large language model (llm)
  hallucination.
\newblock In {\em European Semantic Web Conference}, pages 182--185. Springer,
  2023.

\bibitem{wei2024measuring}
Jiaheng Wei, Yuanshun Yao, Jean-Francois Ton, Hongyi Guo, Andrew Estornell, and
  Yang Liu.
\newblock Measuring and reducing llm hallucination without gold-standard
  answers.
\newblock {\em arXiv preprint arXiv:2402.10412}, 2024.

\bibitem{hanheide2017robot}
Marc Hanheide, Moritz G{\"o}belbecker, Graham~S Horn, Andrzej Pronobis,
  Kristoffer Sj{\"o}{\"o}, Alper Aydemir, Patric Jensfelt, Charles Gretton,
  Richard Dearden, Miroslav Janicek, et~al.
\newblock Robot task planning and explanation in open and uncertain worlds.
\newblock {\em Artificial Intelligence}, 247:119--150, 2017.

\bibitem{jiang2019open}
Yuqian Jiang, Nick Walker, Justin Hart, and Peter Stone.
\newblock Open-world reasoning for service robots.
\newblock In {\em Proceedings of the international conference on automated
  planning and scheduling}, volume~29, pages 725--733, 2019.

\bibitem{zhou2023don}
Kun Zhou, Yutao Zhu, Zhipeng Chen, Wentong Chen, Wayne~Xin Zhao, Xu~Chen,
  Yankai Lin, Ji-Rong Wen, and Jiawei Han.
\newblock Don't make your llm an evaluation benchmark cheater.
\newblock {\em arXiv preprint arXiv:2311.01964}, 2023.

\bibitem{white2024livebench}
Colin White, Samuel Dooley, Manley Roberts, Arka Pal, Ben Feuer, Siddhartha
  Jain, Ravid Shwartz-Ziv, Neel Jain, Khalid Saifullah, Siddartha Naidu, et~al.
\newblock Livebench: A challenging, contamination-free llm benchmark.
\newblock {\em arXiv preprint arXiv:2406.19314}, 2024.

\bibitem{wang2024benchmark}
Siyuan Wang, Zhuohan Long, Zhihao Fan, Zhongyu Wei, and Xuanjing Huang.
\newblock Benchmark self-evolving: A multi-agent framework for dynamic llm
  evaluation.
\newblock {\em arXiv preprint arXiv:2402.11443}, 2024.

\bibitem{wang2023pandalm}
Yidong Wang, Zhuohao Yu, Zhengran Zeng, Linyi Yang, Cunxiang Wang, Hao Chen,
  Chaoya Jiang, Rui Xie, Jindong Wang, Xing Xie, et~al.
\newblock Pandalm: An automatic evaluation benchmark for llm instruction tuning
  optimization.
\newblock {\em arXiv preprint arXiv:2306.05087}, 2023.

\bibitem{ding2025decoupling}
Yongkang Ding, Xiaoyin Wang, Hao Yuan, Meina Qu, and Xiangzhou Jian.
\newblock Decoupling feature-driven and multimodal fusion attention for
  clothing-changing person re-identification.
\newblock {\em Artificial Intelligence Review}, 58(8):241, 2025.

\bibitem{han2025multimodal}
Xiaofeng Han, Shunpeng Chen, Zenghuang Fu, Zhe Feng, Lue Fan, Dong An, Changwei
  Wang, Li~Guo, Weiliang Meng, Xiaopeng Zhang, et~al.
\newblock Multimodal fusion and vision-language models: A survey for robot
  vision.
\newblock {\em arXiv preprint arXiv:2504.02477}, 2025.

\bibitem{chen2022lako}
Zhuo Chen, Yufeng Huang, Jiaoyan Chen, Yuxia Geng, Yin Fang, Jeff~Z Pan, Ningyu
  Zhang, and Wen Zhang.
\newblock Lako: Knowledge-driven visual question answering via late
  knowledge-to-text injection.
\newblock In {\em Proceedings of the 11th International Joint Conference on
  Knowledge Graphs}, pages 20--29, 2022.

\bibitem{xiao2024florence}
Bin Xiao, Haiping Wu, Weijian Xu, Xiyang Dai, Houdong Hu, Yumao Lu, Michael
  Zeng, Ce~Liu, and Lu~Yuan.
\newblock Florence-2: Advancing a unified representation for a variety of
  vision tasks.
\newblock In {\em Proceedings of the IEEE/CVF Conference on Computer Vision and
  Pattern Recognition}, pages 4818--4829, 2024.

\bibitem{lv2022deep}
Zhihan Lv, Fabio Poiesi, Qi~Dong, Jaime Lloret, and Houbing Song.
\newblock Deep learning for intelligent human--computer interaction.
\newblock {\em Applied Sciences}, 12(22):11457, 2022.

\bibitem{zhen2023human}
Rui Zhen, Wenchao Song, Qiang He, Juan Cao, Lei Shi, and Jia Luo.
\newblock Human-computer interaction system: A survey of talking-head
  generation.
\newblock {\em Electronics}, 12(1):218, 2023.

\bibitem{duan2024human}
Haonan Duan, Yifan Yang, Daheng Li, and Peng Wang.
\newblock Human--robot object handover: Recent progress and future direction.
\newblock {\em Biomimetic Intelligence and Robotics}, 4(1):100145, 2024.

\bibitem{yang2024binding}
Fengyu Yang, Chao Feng, Ziyang Chen, Hyoungseob Park, Daniel Wang, Yiming Dou,
  Ziyao Zeng, Xien Chen, Rit Gangopadhyay, Andrew Owens, et~al.
\newblock Binding touch to everything: Learning unified multimodal tactile
  representations.
\newblock In {\em Proceedings of the IEEE/CVF Conference on Computer Vision and
  Pattern Recognition}, pages 26340--26353, 2024.

\bibitem{hao2025tla}
Peng Hao, Chaofan Zhang, Dingzhe Li, Xiaoge Cao, Xiaoshuai Hao, Shaowei Cui,
  and Shuo Wang.
\newblock Tla: Tactile-language-action model for contact-rich manipulation.
\newblock {\em arXiv preprint arXiv:2503.08548}, 2025.

\bibitem{ma2025cltp}
Wenxuan Ma, Xiaoge Cao, Yixiang Zhang, Chaofan Zhang, Shaobo Yang, Peng Hao,
  Bin Fang, Yinghao Cai, Shaowei Cui, and Shuo Wang.
\newblock Cltp: Contrastive language-tactile pre-training for 3d contact
  geometry understanding.
\newblock {\em arXiv preprint arXiv:2505.08194}, 2025.

\end{thebibliography}


%
%

\end{document}